\documentclass[manuscript,screen]{acmart}
\usepackage{dirtytalk}
\usepackage{amsmath}
\usepackage{bbm}
\usepackage{multirow}
\usepackage{makecell}
\usepackage{tikz}
\usepackage{forest}
\usepackage{overpic}
\usepackage{subfigure}
\usepackage{xspace}
\usepackage{caption}
\usepackage{booktabs}
\AtBeginDocument{%
  \providecommand\BibTeX{{%
    \normalfont B\kern-0.5em{\scshape i\kern-0.25em b}\kern-0.8em\TeX}}}

\setcopyright{acmcopyright}
\copyrightyear{2018}
\acmYear{2018}
\acmDOI{XXXXXXX.XXXXXXX}

\acmPrice{15.00}
\acmISBN{978-1-4503-XXXX-X/18/06}

\usetikzlibrary{trees,positioning,shapes,shadows,arrows.meta}

\definecolor{myorange}{RGB}{245, 121, 58}
\definecolor{mypurple}{RGB}{169, 90, 161}
\definecolor{mylightblue}{RGB}{133, 192, 249}
\definecolor{myblue}{RGB}{15, 32, 128}

\begin{document}

\title{Efficient Automation of Neural Network Design: A Survey on Differentiable Neural Architecture Search}


\author{Alexandre Heuillet}
\email{alexandre.heuillet@univ-evry.fr}
\affiliation{%
  \institution{IBISC, Univ Evry, Université Paris-Saclay}
  \country{France}}
\affiliation{
    \institution{Massachusetts Institute of Technology}
    \country{USA}
}
  
\author{Ahmad Nasser}
\email{ahmad.nasser@universite-paris-saclay.fr}
\affiliation{%
  \institution{IBISC, Univ Evry, Université Paris-Saclay}
  \country{France}
}
  
\author{Hichem Arioui}
\email{hichem.arioui@univ-evry.fr}
\affiliation{%
  \institution{IBISC, Univ Evry, Université Paris-Saclay}
  \country{France}}

\author{Hedi Tabia}
\email{hedi.tabia@univ-evry.fr}
\affiliation{%
  \institution{IBISC, Univ Evry, Université Paris-Saclay}
  \country{France}}


\begin{abstract}
  In the past few years, Differentiable Neural Architecture Search (DNAS) rapidly imposed itself as the trending approach to automate the discovery of deep neural network architectures. This rise is mainly due to the popularity of DARTS, one of the first major DNAS methods. In contrast with previous works based on Reinforcement Learning or Evolutionary Algorithms, DNAS is faster by several orders of magnitude and uses fewer computational resources. In this comprehensive survey, we focus specifically on DNAS and review recent approaches in this field. Furthermore, we propose a novel challenge-based taxonomy to classify DNAS methods. We also discuss the contributions brought to DNAS in the past few years and its impact on the global NAS field. Finally, we conclude by giving some insights into future research directions for the DNAS field.
\end{abstract}

\ccsdesc[500]{General and reference~Surveys and overviews}
\ccsdesc[500]{Computing methodologies~Search methodologies}
\ccsdesc[500]{Computing methodologies~Computer vision}

\keywords{deep learning, neural architecture search, meta learning}

\maketitle

\section{Introduction}
Neural Architecture Search (NAS) has witnessed rapid development in recent years. The automation of this field supported the development of novel Deep Learning (DL) \cite{lecun2015deep} architectures, especially Convolutional Neural Networks (CNN) \cite{lecun1995convolutional}, that competed with previous state-of-the-art handcrafted models. Since the introduction of CNNs with LeNet \cite{lecun1995convolutional} and the beginning of Deep Learning with AlexNet \cite{krizhevsky2009learning}, most improvements in the field (e.g., deepening the architecture or adding residual connections) were driven by empiricism. NAS aims to put an end to this trial-and-error practice and bring a formal way to smoothen the progress in deep learning architecture design. Moreover, automatically discovering more efficient architectures is particularly relevant in the ecological transition context (i.e., green Deep Learning \cite{xu2021survey}). The role of manual feature engineering and model development has gradually decreased ever since. Consequently, new challenges such as memory efficiency, transferability between datasets, or computational efficiency have been introduced in the field. This has encouraged researchers to integrate all possible approaches in the literature to improve NAS methodologies. Notably, parameter space differentiability, which uses state-of-the-art optimizers for deep learning model training, is considered one of the most promising leads.

\begin{figure}[h]
  \centering
  \includegraphics[width=0.7\linewidth]{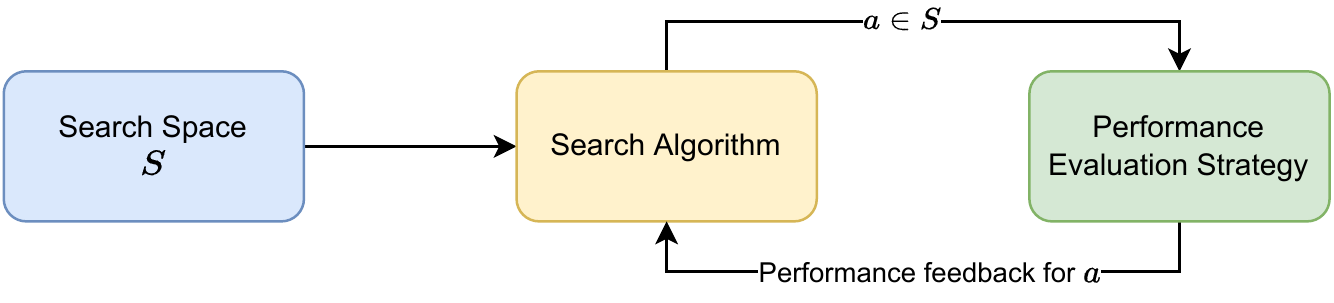}
  \caption{Typical layout of a Neural Architecture Search framework.}
  \label{fig:nas_components}
\end{figure}

A typical NAS method always consists of three components regardless of the methodology employed, as visually summarized in Fig \ref{fig:nas_components}. Firstly, a search space that comprises all the possible hyperparameter sets of the architecture. Generally, this is a discrete search space as it often only covers the categorical choice of operations that compose the architecture. However, it can be made continuous by relaxing this choice using a projection function (e.g., \textit{softmax}) \cite{liu2018darts} or by including continuous hyperparameters (e.g., the learning rate). It is theoretically infinite, but in practice, it is limited to finite bounds to reduce computational cost and avoid too deep/dense/big architectures. Secondly, a search strategy also denoted as search algorithm or optimizer, that is responsible for exploring the search space and sampling candidate architectures by taking into account the performance feedback of candidates previously sampled. Finally, an evaluation strategy that evaluates each candidate architecture selected by the search strategy. This strategy aims to estimate the model's performance, ideally without the need to fully train the model to reduce computational cost.

In the literature, handcrafted approaches have first been used to find optimal deep learning architectures. However, in the past few years, ML-based methodologies emerged to accurately address the NAS problem within well-defined constraints. In the following subsections, we briefly present the main categories of NAS approaches as well as the organization of this article.


\subsection{Tabular-based Approaches}

 The first methods that were implemented to automate neural network design were heuristic-based approaches that process tabular data. Such methods include random search \cite{li2020random}, regular grid sampling of search space \cite{geifman2019deep}, tree-based sampling \cite{jin2019auto}, and many other methods inspired by the field of optimization \cite{xie2018snas, hu2020dsnas, kandasamy2018neural, liu2018progressive}. Although quickly eclipsed by ML-based approaches (see the following subsections) and especially Differentiable NAS, these methods remain present in the NAS landscape as several ones have been proposed in the past few years \cite{hu2020dsnas, li2020random}.

\subsection{Reinforcement Learning}

Zoph et al. \cite{zoph2016neural} deployed Reinforcement Learning (RL) to drive the architecture search process. This was one of the first ML-based NAS methods. The RL controller (or agent) continuously chooses new sets of hyperparameters based on the previously evaluated ones. For each set of hyperparameters, the model is fully trained on a given dataset, and a final evaluation score is used as a reward for the RL controller. Since fully training each candidate model is time-consuming, the authors implemented parallelism and asynchronous parameter update to speed up the search process. Later on, Zoph et al. \cite{zoph2018learning} introduced the concept of cell-based hyperparameter tuning, where only the internal architecture of a building block (denoted “cell”) is optimized instead of a complete neural network. Thus, the final model consists of a stacked series of cells. This structural approach, called NASNet, was employed to support model transferability and reduce the search space size. Although NASNet has offered new advantages to help with model optimization, it imposed new challenges concerning the gap between the developing cell structure and the actual multi-cell model derived for evaluation. This structural concept has later migrated to other NAS approaches, as discussed later.

\subsection{Evolutionary Algorithms}

Evolutionary Algorithms (EA) have also presented a valid approach for NAS. Applying EA to NAS is straightforward, as hyperparameter sets are \say{genotypes} that define an architecture. In EA, a Darwinist process gradually improves an initial population of randomly initialized architectures over several generations. A new generation is obtained by deriving \say{children} through the recombination (crossover) of genes from the best individuals in the current population. Recent works implemented evolutionary strategies such as guided evolution \cite{lopes2021guided}, reinforced evolution \cite{chen2019renas}, and regularized EA \cite{real2019regularized}. These NAS approaches were among the first to be implemented. However, they are computationally expensive, especially when handling vast search spaces, as they blindly explore the search space during the first iterations. Thus, the overall computational time in GPU days is often ludicrous (e.g., more than 3000 GPU days for AmoebaNet \cite{real2019regularized}). Moreover, most works in the literature are case-specific. Hence, transferability was less practical than in later gradient-based approaches.

\subsection{Gradient Descent and Differentiability}

Gradient descent is a powerful technique that has been known since the early 19th century with the works of French mathematician Augustin-Louis Cauchy \cite{lemarechal2012cauchy}. Its practical implementation is nearly as old as computer science itself, as it was first explored by Haskell Curry in the 1940s \cite{curry1944method}. Since then, it became increasingly well-studied and eventually led to the tremendous success of ML with the gradient backpropagation algorithm \cite{rumelhart1986learning}. More specifically, gradient-based methods for hyperparameter optimization have been in the literature since the early 1990s \cite{arai1991adaptive, chen1996feedforward}. Similar to many contributions to Machine Learning at the time, these works were responding to control challenges. They implemented gradient descent for automatically selecting and changing the shape of activation functions in artificial neural networks. In a practical sense, this process is similar to weighing the neurons of a neural network, which affects the model’s output shape. The past few years saw the rise of a novel approach for combining gradient descent and NAS: Differentiable NAS (DNAS). 

DNAS implements gradient-based methods to tune the hyperparameters of deep learning models as first pointed out by Bender et al. \cite{bender2018understanding}. The hyperparameter tuning problem is converted into a continuous optimization problem by considering the search space as a smooth manifold. Similar to model training, DNAS uses the gradient information of hyperparameter weights to find the optimal set. This makes it possible to use a supernetwork that instantiates in memory all candidate architectures, thus removing the need to evaluate each candidate independently to get performance feedback. As a result, fewer computational resources are used to reach the optimal set compared to other implementations, such as evolutionary strategies or reinforcement learning. As shown earlier, implementing such approaches requires a relaxation of the discrete search space. This has been first achieved by Differentiable ARchitecTure Search (DARTS) \cite{liu2018darts} using a cell-based paradigm for the search space. From this point onward, DARTS became increasingly popular, helping to democratize NAS with its computational efficiency.

Importantly, DNAS and the other ML-based NAS methods discussed above are generally considered black-box approaches. However, ML explainability (eXplainable Artificial Intelligence, XAI) is starting to emerge \cite{lundberg2017unified, heuillet2021explainability}. This point will be more thoroughly discussed in Section \ref{sec:discussion}.

\subsection{Article Organization}
In this survey, we focused on Differentiable Neural Architecture Search (DNAS) as a trending approach to perform NAS. We analyzed 26 papers that have appeared since 2019 in leading machine learning and computer vision conferences and journals. The goal
is to help the reader navigate this emerging field, which gained significant momentum in the past few years. To the extent of our knowledge, we are the first to propose a survey specifically centered on DNAS. We also put the emphasis on computer vision tasks, although other fields of application are discussed in Section \ref{sec:applications}. Compared with the existing survey literature \cite{xie2021weight, ren2021comprehensive}, the main contributions of this article are as follows:

\begin{itemize}
    \item An attempt to create a taxonomy of DNAS (Section \ref{sec:taxonomy}) where we identify the main challenges posed by DNAS (Section \ref{sec:darts}).
    \item A comprehensive review of recent DNAS works according to the challenges they addressed (Section \ref{sec:review}). We provide a summary of the reviewed methods in tabular form (Table \ref{tab:applications}).
    \item A discussion on the impact of DNAS on NAS and Deep Learning, along with information on Explainable NAS and insights on future research directions (Section \ref{sec:discussion}). 
\end{itemize}

Finally, in Section \ref{sec:conclusion}, we bring a conclusion to this article.

\section{Problem statement and taxonomy of differentiable NAS}
\label{sec:taxonomy}
This survey article aims to discuss and analyze Differentiable NAS works according to a novel taxonomy. DNAS methods are typically categorized into two classes: \textbf{(a)} DARTS \cite{liu2018darts} derivatives, to which the majority (62 \%) of prior works we reviewed belong, and \textbf{(b)} all other DNAS studies. Works in \textbf{(a)} assumed that DARTS is on the right track and could be further improved to go past its limitations (detailed in Section \ref{sec:darts}). The fact that a large number of works belong to \textbf{(a)} can be explained by the high popularity that DARTS enjoyed in the past few years. In 2022, DARTS was already an old method (having been published in 2019) but continued to inspire new publications \cite{ye2022b, yan2022radars}, thus demonstrating it passed the test of time. Studies in \textbf{(b)} proposed novel DNAS algorithms and search spaces. In many cases \cite{wu2019fbnet, cai2018proxylessnas, wan2020fbnetv2}, they would assume that DARTS' issues are inherent to its conception and that a \textit{tabula rasa} approach was necessary.

\begin{figure}[H]
 \centering
 \small
 \tikzset{
    basic/.style  = {draw, text width=3cm, align=center, font=\sffamily, rectangle},
    root/.style   = {basic, rounded corners=2pt, thin, align=center, fill=green!30},
    onode/.style = {basic, thin, rounded corners=2pt, align=center, fill=green!60,text width=3cm,},
    tnode/.style = {basic, thin, fill=myorange!60, text width=16em, align=center},
    bnode/.style = {basic, thin, fill=mylightblue!60, text width=16em, align=center},
    xnode/.style = {basic, thin, rounded corners=2pt, align=center, fill=red!20,text width=4cm,},
    wnode/.style = {basic, thin, align=left, fill=pink!10!blue!80!red!10, text width=6.5em},
    edge from parent/.style={draw=black, edge from parent fork right}

}
 \begin{forest} for tree={
    grow=east,
    growth parent anchor=west,
    parent anchor=east,
    child anchor=west,
    edge path={\noexpand\path[\forestoption{edge},->, >={latex}] 
         (!u.parent anchor) -- +(10pt,0pt) |-  (.child anchor) 
         \forestoption{edge label};}
}
[Differentiable Neural Architecture Search, root, s sep=0.6cm,  l sep=10mm,
    [Search Space Restrictions, xnode, s sep=0.5cm,  l sep=10mm,
        [D-DARTS\cite{heuillet2021d}\, DenseNAS\cite{fang2020densely}, tnode]
        [ProxylessNAS\cite{cai2018proxylessnas}\, FBNet\cite{wan2020fbnetv2}\, FBNetV2\cite{wu2021fbnetv5}\, FBNetV5\cite{wu2021fbnetv5}, bnode] ]
    [Computational Efficiency and Latency Reduction, xnode, s sep=0.5cm,  l sep=10mm,
        [PC-DARTS\cite{xu2019pc}\, DOTS\cite{gu2021dots}, tnode]
        [ProxylessNAS\cite{cai2018proxylessnas}\, FBNet\cite{wu2019fbnet}\, FBNetV2 \cite{wan2020fbnetv2}\, VIM-NAS \cite{wang2021learning}\, HardCoRe-NAS\cite{nayman2021hardcore}\, RADARS\cite{yan2022radars}, bnode] ]
    [Over-representation of \textit{skip connections}, xnode, s sep=0.5cm,  l sep=10mm,
        [FairDARTS\cite{chu2020fair}\, P-DARTS\cite{chen2019progressive}\, DARTS+\cite{liang2019darts+}\, R-DARTS\cite{Zela2020Understanding}\, DARTS-\cite{chu2020darts}\, PR-DARTS\cite{zhou2020theory}\, DARTS+PT\cite{wang2021rethinking}\, NoisyDARTS\cite{chunoisy2021}\, $\beta$-DARTS\cite{ye2022b}\, CDARTS\cite{yu2022cyclic}, tnode]
        [$\text{E}^2\text{NAS}$\cite{zhang2020differentiable}, bnode]]
    [Gradient Approximation Inconsistencies and Optimization Gap, xnode, s sep=0.5cm,  l sep=10mm,
        [FairDARTS\cite{chu2020fair}\, P-DARTS\cite{chen2019progressive}\, S-DARTS\cite{chen2020stabilizing}\, iDARTS\cite{zhang2021idarts}\, DARTS+PT\cite{wang2021rethinking}\, $\beta$-DARTS\cite{ye2022b}\, DOTS\cite{gu2021dots}\, CDARTS\cite{yu2022cyclic}, tnode]
        [ProxylessNAS\cite{cai2018proxylessnas}\, UNAS\cite{Vahdat_2020_CVPR}\, EnTranNAS\cite{yang2021towards}\, $\text{E}^2\text{NAS}$\cite{zhang2020differentiable}\, OFA\cite{cai2020once}\, BigNAS\cite{yu2020bignas}, bnode] ] ]
\end{forest}
\caption{
Taxonomy of the reviewed Differentiable Neural Architecture Search literature. References in \textcolor{myorange}{orange}, and \textcolor{mylightblue}{light blue} correspond to \textcolor{myorange}{DARTS-based}, and \textcolor{mylightblue}{non-DARTS-based} approaches respectively.
}
\label{fig:taxonomic_tree}
\end{figure}
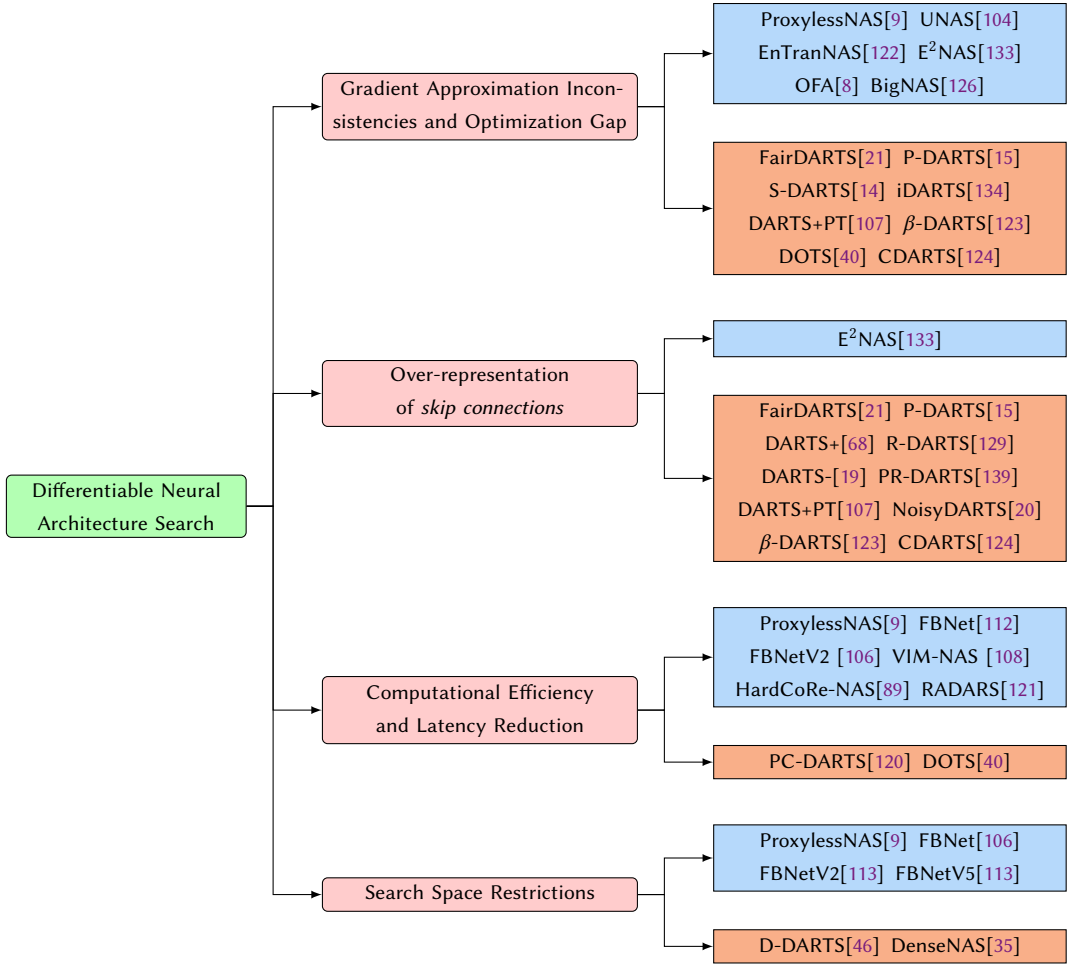

However, this classification is trivial and does not help navigate the DNAS landscape. Thus, in this article, we propose a novel taxonomy where methods are classified according to the challenge they attempt to solve rather than belonging to \textbf{(a)} or \textbf{(b)}. We identified 4 different challenges: \textbf{(I)} Bridging the optimization gap between the proxy (i.e., used during search) and final models and resolving gradient approximation issues, \textbf{(II)} Countering the over-representation of non-parametric operations (e.g., \textit{skip connections}, \textbf{(III)} Improving computational efficiency and reducing latency at inference time, and \textbf{(IV)} Bypassing the search space restrictions inherent to DNAS. These challenges are presented in detail in Section \ref{sec:darts}. Fig. \ref{fig:taxonomic_tree} presents a taxonomic tree summarizing the different works we reviewed and our proposed DNAS classification.

\section{DARTS and the challenges of Differentiable NAS}
\label{sec:darts}
In this section, we present the Differentiable ARchiTecture Search (DARTS) family, which is currently the most popular DNAS family of approaches, and we analyze its limitations along with the development it has inspired in the literature.

DARTS was introduced by Liu et al. \cite{liu2018darts} as a novel method to implement DNAS with a cell-based approach. In contrast to the majority of evolutionary algorithms \cite{van2019evolutionary, chen2019renas, lopes2021guided, real2019regularized}, and other DNAS methods such as FBNet \cite{wu2019fbnet, wan2020fbnetv2, wu2021fbnetv5}, DARTS defines a modular search space composed of a few building blocks called \say{cells}, similar to the one used in some RL-based works \cite{zoph2016neural, zoph2018learning}. There are two types of cells: normal cells (i.e., that make up most of the architecture) and reduction cells (i.e., that perform dimension reduction). Each cell is a direct acyclic graph comporting $N$ nodes (i.e., intermediary representations of data such as feature maps) linked by edges. Each edge $e_{i,j}$ connecting node $i$ to node $j$ is the sum of the outputs of $|O_{i,j}|=K$ operations, where $O_{i,j} = \{o^{1}_{i,j}, ..., o^{K}_{i,j}\}$ represents the set of all possible operations for $e_{i,j}$. As a result, each node receives a combination of operation outputs from all previous nodes. When searching for CNN cells, the search space of DARTS is composed of $K=7$ operations: \textit{skip\_connect}, \textit{max\_pool\_3x3}, \textit{avg\_pool\_3x3}, \textit{sep\_conv\_3x3}, \textit{sep\_conv\_5x5}, \textit{dil\_conv\_3x3} and \textit{dil\_conv\_5x5}.

\begin{figure}[h]
  \centering
  \includegraphics[width=0.7\linewidth]{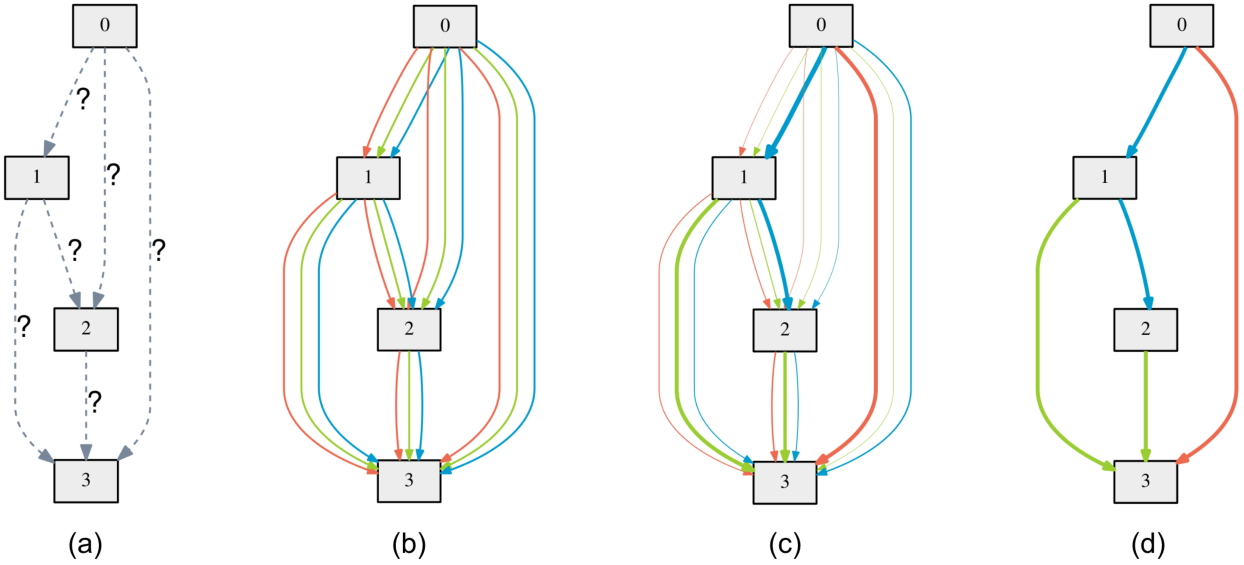}
  \caption{Overview of the search strategy of DARTS. Figure from Liu et al. \cite{liu2018darts}.}
  \label{fig:darts_pruning}
\end{figure}

Hence, DARTS' search strategy is to progressively prune incoming edges from each cell node until a maximum of 2 remains (see Fig \ref{fig:darts_pruning}). To that end, each operation $o \in O_{i,j}$ in the mixed output of each edge $e_{i,j}$ is associated to a specific weight $\alpha^{k}_{i,j} \in \alpha_{i,j}= \{\alpha^{1}_{i,j}, ..., \alpha^{K}_{i,j}\}$ with $\alpha_{i,j} \in \alpha$. DARTS relaxes the categorical choice of operations for $e_{i,j}$ into a continuous form (i.e., a probability distribution) by performing a \textit{softmax} operation on $\alpha_{i,j}$. Thus, the mixed output $\overline{o}_{i,j}$ of $e_{i,j}$ is defined as
\begin{equation}
    \overline{o}_{i,j}(x) = \sum^{K}_{k=1} \frac{exp(\alpha^{k}_{i,j})}{\sum^{K}_{k'=1} exp(\alpha^{k'}_{i,j})}o^{k}_{i,j}(x) = \sum^{K}_{k=1} \text{softmax}(\alpha^{k}_{i,j})o^{k}_{i,j}(x),
\label{eq:mixed_output}
\end{equation}
where $x$ is the input feature and $\alpha^{k}_{i,j} \in \alpha_{i,j}$ is the weight associated with operation $o^k_{i,j} \in O_{i,j}$.

The architectural parameters $\alpha$ are learned through gradient descent to minimize the validation loss $\mathcal{L}_{val}$ while a supernet (also denoted proxy network) is trained to minimize the training loss $\mathcal{L}_{train}$. This supernet comprises a small number (e.g., 8) of stacked search cells and is a representation of all candidate architectures. This way, DARTS solves a bi-level optimization problem formulated as 
\begin{equation}
    \begin{split}
        \underset{\alpha}{\text{min }} \mathcal{L}_{val}(w^{*}(\alpha),\alpha),\\
    \text{s.t.} w^{*}(\alpha) = \underset{w}{\text{argmin }} \mathcal{L}_{train}(w,\alpha),
    \end{split}
\label{eq:bilevel_optimization}
\end{equation}
where $w$ denotes the supernet weights.
The gradient of the architectural parameters $\alpha$ is thus computed as follows:
\begin{equation}
    \Delta_{\alpha}\mathcal{L}_{val} = \frac{\partial \mathcal{L}_{val}}{\partial \alpha} + \frac{\partial \mathcal{L}_{val}}{\partial w} \frac{\partial w^{*}(\alpha)}{\partial \alpha},
    \label{eq:gradient_calc}
\end{equation}
where $w^{*}(\alpha) = \text{argmin}_w \mathcal{L}_{train}(w,\alpha)$ (i.e., the optimal value of $w$ obtained by minimizing the training loss $\mathcal{L}_{train}$).
Finally, once the search phase is over, the internal structure of each type of cell is discretized into the final model by edge selection through a \textit{softmax}. It is possible to derive a final architecture of any size by simply stacking repetitive sequences of the two types of cells. 

This way DARTS' search process does not need to fully train each candidate architecture and instead use the supernet as an approximator to get performance feedback by considering the current weights $w$ as equivalent to the optimal weights $w^*$. Hence, DARTS is significantly faster than RL-based or EA-based approaches.

However, DARTS suffers from major limitations and poses new challenges. Firstly, as discussed by several articles \cite{heuillet2021d, chu2020darts, chu2020fair, ye2022b}, gradient approximation methods inevitably cause \textbf{inconsistencies in the optimization process (I)} that affect the architectural parameters. Combined with the limited convergence ability of \textit{softmax} (itself a smooth approximation of the \textit{argmax} function), this leads to a final probability distribution mainly composed of values very close to one another (i.e., with a low standard deviation compared to the mean). Having only a small difference between the highest values of the probability distribution makes the discretization process more challenging. For instance, it is non-trivial to assert that operation $o_1$ associated with probability $p_{o_1}=0.92$ should be selected rather than operation $o_2$ associated with probability $p_{o_2}=0.91$. Hence, it is noteworthy to point out that there is a significant gap between the proxy network (smaller, with mixed outputs on edges) used during the search process and the final discretized model (larger, with a maximum of 2 operations per edge). Chen et al. \cite{chen2019progressive} highlighted that this gap also appears when transferring the model to a different dataset than the one used during the search phase. 

\begin{figure}[h]
  \centering
  \includegraphics[width=\linewidth]{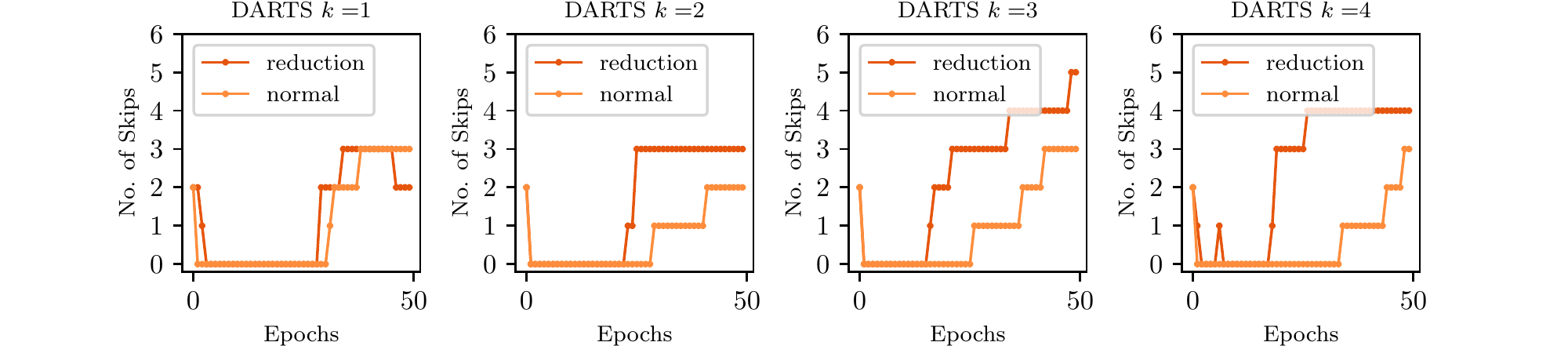}
  \caption{Evolution of the number of \textit{skip connections} w.r.t. the number of search epochs. Figure from Chu et al. \cite{chu2020fair}}
  \label{fig:skip_connections}
\end{figure}

Secondly, DARTS tends to \textbf{overly represent \textit{skip connections} (II)} within the discovered architectures. This problem is related to the first one since, as argued by Chu et al. \cite{chu2020fair}, the \textit{softmax} operation increases exclusive competition between the different operations (i.e., if one operation is favored, it is at the cost of the others). Edges that include at least one \textit{skip connection} resemble residual blocks \cite{he2016deep} and quickly provide a performance boost (i.e., by accelerating forward and backward operations) over other operations that are hence suppressed by \textit{softmax}. In addition, \textit{skip connections} are unparameterized (weight-free) operations and thus have a limited ability to learn data representations. This leads to a global performance collapse as \textit{skip connections} are selected even in edges where they are not the fittest operation. Fig \ref{fig:skip_connections} showcases an illustration of the \textit{skip connections} over-representation phenomenon. 

Thirdly, another challenge posed by DARTS is the ability to \textbf{efficiently browse the search space (III)} (i.e., to limit the amount of computational resources used by the search algorithm). Eq. \ref{eq:mixed_output} shows that every possible path connecting every pair of nodes together must be instantiated in memory in the form of a supernet. Although limited when using toy/proxy datasets such as CIFAR-10 and CIFAR-100 \cite{krizhevsky2009learning}, this is especially concerning when using large real-world-like datasets such as ImageNet \cite{deng2009imagenet} or MS-COCO \cite{lin2014microsoft}. With modern Nvidia GPUs (i.e., starting from Volta), it is possible to take advantage of Tensor Cores (built-in hardware specialized in matrix multiplication) by using Automatic Mixed Precision (AMP) \cite{micikevicius2018mixed}. This training process uses IEEE half-precision format (FP16) \cite{ieee2019precision} when single-precision (FP32) is not needed, hence speeding up computations and resulting in a smaller memory footprint. However, using a lower precision can sometimes cause numerical instability (e.g., gradient overflow). Thus, researchers should primarily focus on designing efficient DNAS approaches instead of only relying on AMP.

Finally, \textbf{DARTS' search space is very restricted (IV)}. Searching only for two building blocks leads to a substantially simpler optimization problem. Still, it is insufficient as empirical evidence showed that most modern high-performing CNN architectures such as ResNets \cite{he2016deep}, ResNext \cite{xie2017aggregated}, or EfficientNetV2 \cite{tan2021efficientnetv2} are composed of more than two different blocks. Thus, increasing diversity in the discoverable architectures is one of the challenges that DARTS-based methods must overcome.

In Section \ref{sec:review}, we reviewed some of the most impactful DNAS methods (DARTS derivatives and other methods) and analyzed how they address the four challenges we identified \textbf{(I, II, III, and IV)}.

\section{Literature Review of Differentiable NAS}
\label{sec:review}
In this section, we perform a comprehensive review of recent DNAS literature, with a focus on its most studied subcategory: DARTS \cite{liu2018darts} and its derivatives. Each of the \textbf{28 approaches} we reviewed is presented in the following subsections according to the challenge (see Section \ref{sec:darts}) it attempted to solve, according to our proposed taxonomy presented in Section \ref{sec:taxonomy}. Some approaches are listed in several subsections as they address multiple DNAS challenges. We present a summary of this review in Table \ref{tab:review}.

\begin{table}[h]
    \centering
    \small
    \caption{Summary of the literature we reviewed in this survey. We included DARTS \cite{liu2018darts} for comparison purposes. The search cost is expressed with the GPU used by the authors of the original article. $\dagger$ denotes models searched on CIFAR-10 or CIFAR-100 \cite{krizhevsky2009learning}.}
    \begin{tabular}{llllll}
        \toprule
        \thead{Title} & \thead{Type} & \thead{Challenges\\ tackled} & \thead{Top 1 accuracy\\
        on ImageNet (\%)} & \thead{Search cost\\ (GPU days)} & \thead{Search space} \\
        \midrule
         DARTS \cite{liu2018darts} & N.A. & N.A. & 73.1 & 4 & \textit{darts}\\
         ProxylessNAS \cite{cai2018proxylessnas} & other & \textbf{(I, III, IV)} & 75.1 & 8.3 & \textit{MobileNet-like}\\
         P-DARTS$\dagger$ \cite{chen2019progressive} & DARTS-based & \textbf{(I, II)} & 75.3 & 0.3 & \textit{darts}\\
         FBNet \cite{wu2019fbnet} & other & \textbf{(III, IV)} & 74.9 & 9 & \textit{MobileNet-like}\\
         PC-DARTS \cite{xu2019pc} & DARTS-based & \textbf{(III)} & 75.8 & 3.8 & \textit{darts}\\
         DARTS+ \cite{liang2019darts+} & DARTS-based & \textbf{(II)} & 76.1 & 6.8 & \textit{MobileNet-like}\\
         PR-DARTS$\dagger$ \cite{zhou2020theory} & DARTS-based & \textbf{(II)} & 75.9 & 0.17 & \textit{darts}\\
         OFA \cite{cai2020once} & other & \textbf{(I, III)} & 80.0 & 1.7 & \textit{MobileNet-like}\\
         BigNAS \cite{yu2020bignas} & other & \textbf{(I, III)} & 80.9 & 1.6 & \textit{MobileNet-like}\\
         FBNetV2 \cite{wan2020fbnetv2} & other & \textbf{(III, IV)} & 77.2 & 25 & \textit{custom}\\
         R-DARTS$\dagger$ \cite{Zela2020Understanding} & DARTS-based & \textbf{(II)} & - & 1.6 & \textit{custom}\\
         S-DARTS$\dagger$ \cite{chen2020stabilizing} & DARTS-based & \textbf{(I)} & 74.8 & 1.3 & \textit{darts}\\
         FairDARTS \cite{chu2020fair} & DARTS-based & \textbf{(I, II)} & 75.6 & 3 & \textit{MobileNet-like}\\
         DenseNAS \cite{fang2020densely} & DARTS-based & \textbf{(IV)} & 75.3 & 2.7 & \textit{MobileNet-like}\\
         UNAS \cite{Vahdat_2020_CVPR} & other & \textbf{(I)} & 75.5 & 4.3 & \textit{MobileNet-like}\\ 
         iDARTS$\dagger$ \cite{zhang2021differentiable} & DARTS-based & \textbf{(I)} & 75.7 & - & \textit{darts}\\
         FBNetV5 \cite{wu2021fbnetv5} & other & \textbf{(IV)} & 81.7 & >100 & \textit{custom}\\
         NoisyDARTS \cite{chunoisy2021} & DARTS-based & \textbf{(II)} & 76.1 & 12 & \textit{MobileNet-like}\\
         DARTS- \cite{chu2020darts} & DARTS-based & \textbf{(II)} & 76.2 & 4.5 & \textit{MobileNet-like}\\
         VIM-NAS \cite{wang2021learning} & other & \textbf{(III)} & 76.2 & 0.26 & \textit{darts}\\
         DOTS \cite{gu2021dots} & DARTS-based & \textbf{(I, III)} & 76.0 & 1.3 & \textit{darts}\\
         DARTS+PT$\dagger$ \cite{wang2021rethinking} & DARTS-based & \textbf{(I, II)} & 74.5 & 0.8 & \textit{darts}\\
         HardCoRe-NAS \cite{nayman2021hardcore} & other & \textbf{(III)} & 77.9 & 16.7 & \textit{custom}\\
         D-DARTS \cite{heuillet2021d} & DARTS-based & \textbf{(IV)} & 77.0 & 0.3 & \textit{custom}\\
         EnTranNAS \cite{yang2021towards} & other & \textbf{(I)} & 75.7 & 1.9 & \textit{darts}\\
         RADARS \cite{yan2022radars} & other & \textbf{III} & 73.8 & 3.1 & \textit{MobileNet-like}\\
         $beta$-DARTS \cite{ye2022b} & DARTS-based & \textbf{(I, II)} & 76.1 & 0.4 & \textit{darts}\\
         CDARTS \cite{yu2022cyclic} & DARTS-based & \textbf{(I, II)} & 76.3 & 1.7 & \textit{darts}\\
         \bottomrule
    \end{tabular}
    \label{tab:review}
\end{table}

\subsection{Gradient Approximation Inconsistencies and Optimization Gap}
\label{sec:gradient_approximation}
\textbf{FairDARTS} \cite{chu2020fair} tackled DARTS' gradient-related issues by replacing \textit{softmax} for the categorical choice of operations by the \textit{sigmoid} operation $\sigma$. This change is motivated by the fact that, contrary to \textit{softmax}, $\sigma$ does not create exclusive competition between the different operations (i.e., the weights associated with operations can independently increase or decrease). This improves fairness in the operation selection and thus results in better gradient approximations (see Section \ref{sec:gradient_approximation}).In addition, FairDARTS introduced a novel loss function dubbed zero-one loss and denoted $\mathcal{L}_{01}$. This loss function aims to push the sigmoid values of the architectural weights (i.e., $\sigma(\alpha)$) towards 0 or 1 to minimize the discretization gap. Its gradient magnitude is adequately designed to let the $\alpha$ weights fluctuate but still pull them towards 0 or 1 if they stray away from 0.5. $\mathcal{L}_{01}$ is designed as follows:
\begin{equation}
    \mathcal{L}_{01}(\alpha) = -\frac{1}{N}\sum_{i}^{N}(\sigma(\alpha_i)-0.5)^2, 
\end{equation}
where N is the number of nodes in a cell. $\mathcal{L}_{01}$ is differentiable and thus can be backpropagated to help optimize the architectural weights $\alpha$. This loss is combined with $\mathcal{L}_{val}$ to form the total loss
\begin{equation}
    \mathcal{L}_T(w^*,\alpha) = \mathcal{L}_{val}(w^*(\alpha),\alpha) + w_{01}\mathcal{L}_{01}
\end{equation}
where $w_{01}$ is an hyperparameter weighting $\mathcal{L}_{01}$.

\textbf{ProxylessNAS} \cite{cai2018proxylessnas} also sought to close the optimization gap but strayed away from what had been set up by DARTS. The authors proposed to search directly on the target large-scale dataset (e.g., ImageNet \cite{deng2009imagenet}) instead of transferring from a small-scale proxy dataset (e.g., CIFAR-10) as done by DARTS and most of its derivatives. To achieve this, they designed a search space that is no longer composed of repetitive building blocks but instead comprises an entire architecture and includes additional candidate operations. However, this comes at the cost of a greatly increased memory consumption as, if we recall Eq. \ref{eq:mixed_output}, every output feature vector associated with every path in the mixed output of every cell edge must be instantiated and stored in GPU memory. To alleviate this issue, ProxylessNAS replaced DARTS' real-value architectural weights $\alpha$ with binary gates $g$ that output one-hot vectors according to a probability distribution $\{p_1, ..., p_K\}$:
\begin{equation}
    g = \text{binarize}(p_1, ..., p_K) = \left \{ \begin{array}{rl}
    \left[1, 0, ..., 0\right]     & \text{with probability } p_{1}  \\
    ... & \\
    \left[0, 0, ..., 1\right]     & \text{with probability } p_{K} \\ 
    \end{array}
    \right .
\end{equation}
Thus, Eq. \ref{eq:mixed_output} is modified as follows:
\begin{equation}
    \overline{o}_{i,j}^{Binary}(x) = \sum^{K}_{k=1} g_{i,j}^{k}o_{i,j}^{k}(x) = \left \{ \begin{array}{rl}
    o_{i,j}^{1}(x)     & \text{with probability } p_{1}  \\
    ... & \\
    o_{i,j}^{K}(x)     & \text{with probability } p_{K} \\ 
    \end{array}
    \right .
\end{equation}
This mechanism allows entire paths to be binarized and instantiates only one path at a time in memory during the search phase. Thus, it reduces memory consumption to the same level as a regular model. ProxylessNAS successfully overperforms DARTS by 2 \% on ImageNet. Nevertheless, these positive results come with a drastically increased search cost from 1.5 GPU days (DARTS) to 8.3 GPU days.

\begin{figure}[h]
  \centering
  \includegraphics[width=0.5\linewidth]{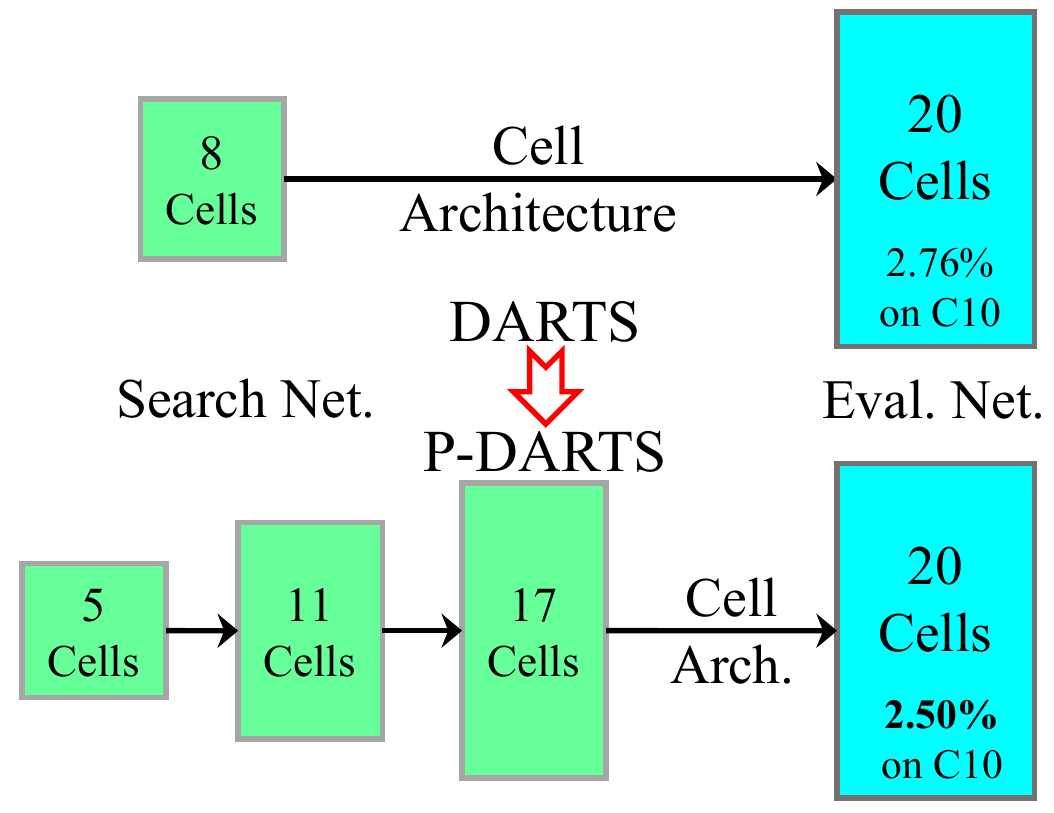}
  \caption{Layout of the P-DARTS search process. Figure from Chen et al. \cite{chen2019progressive}.}
  \label{fig:pdarts}
\end{figure}

Another work, entitled Progressive DARTS (\textbf{P-DARTS}) \cite{chen2019progressive}, focused on reducing the optimization gap between the search and final architectures by improving the proxy model used during search. More precisely, the authors gradually increased the proxy network depth during search (e.g., from 5 cells to 20 cells) in contrast with the original DARTS that uses a fixed 8-cell proxy network that is later derived into a 20-cell final model. Furthermore, the number of candidate operations is progressively reduced according to their performance score. This search space approximation method alleviates the computational efficiency issues encountered when increasing the depth of the proxy network. This process, resumed in Fig. \ref{fig:pdarts}, improved performance by around 0.5 \% on CIFAR-10/CIFAR-100 \cite{krizhevsky2009learning} and reduced the search cost from 1.5 GPU days to 0.3 GPU day compared to DARTS.

SmoothDARTS (\textbf{SDARTS}) was designed by Chen et al. \cite{chen2020stabilizing} as a way to stabilize the bi-level optimization in DARTS (see Eq. \ref{eq:bilevel_optimization}). Similarly to the authors of R-DARTS \cite{Zela2020Understanding} (discussed in Section \ref{sec:skip_connections}), they argued that the optimization gap between the proxy model and the final discretized architecture is highly correlated (inversely proportional) to the spectral norm of the Hessian matrix of the validation loss $\Delta^{2}_{\alpha} \mathcal{L}_{val}$. Hence, they proposed to smooth the validation landscape of DARTS by computing $\underset{\alpha}{\text{min}}\mathcal{L}_{val}(w^*(\alpha),\alpha)$ using $w^*(\alpha)$ obtained either through random smoothing (SDARTS-RS) or through adversarial training (SDARTS-ADV). SDARTS-RS reformulates Eq. \ref{eq:bilevel_optimization} as
\begin{equation}
    w^*(\alpha) = \underset{w}{\text{argmin }} \mathbb{E}_{\delta ~ U_{[-\epsilon, \epsilon]}}\mathcal{L}_{train}(w, \alpha+\delta),
\end{equation}
where $\delta$ is a perturbation sampled from the uniform distribution $U_{[-\epsilon, \epsilon]}$ between $-\epsilon$ and $\epsilon$. The idea behind this is to minimize $\mathcal{L}_{val}(\alpha)$ under a small randomized perturbation $\epsilon$. Similarly, SDARTS-ADV is formulated as follows:
\begin{equation}
    w^*(\alpha) = \underset{w}{\text{argmin }} \underset{||\delta|| < \epsilon}{\text{ max}}\mathcal{L}_{train}(w, \alpha+\delta).
\end{equation}
Here, Chen et al. strove to increase adversarial robustness by minimizing the worst-case loss under a certain perturbation (computed using a multistep Projected Gradient Descent). They theoretically and empirically demonstrated that both SDARTS-RS and SDARTS-ADV improve the stability and generability of DARTS (e.g., SDARTS-ADV overperforms DARTS by +1.1 \% top 1 accuracy on ImageNet). However, both methods have downsides: SDARTS-ADV increases computational cost sharply, but SDARTS-RS is less accurate. Fig. \ref{fig:sdarts} provides an illustration of the smoothing at work in SDARTS.

\def\name{SDARTS\xspace}
\def\EXP{SDARTS-RS\xspace}
\def\ADV{SDARTS-ADV\xspace}

\begin{figure}[h]
  \centering
  \subfigure[DARTS]{\includegraphics[width=0.33\linewidth]{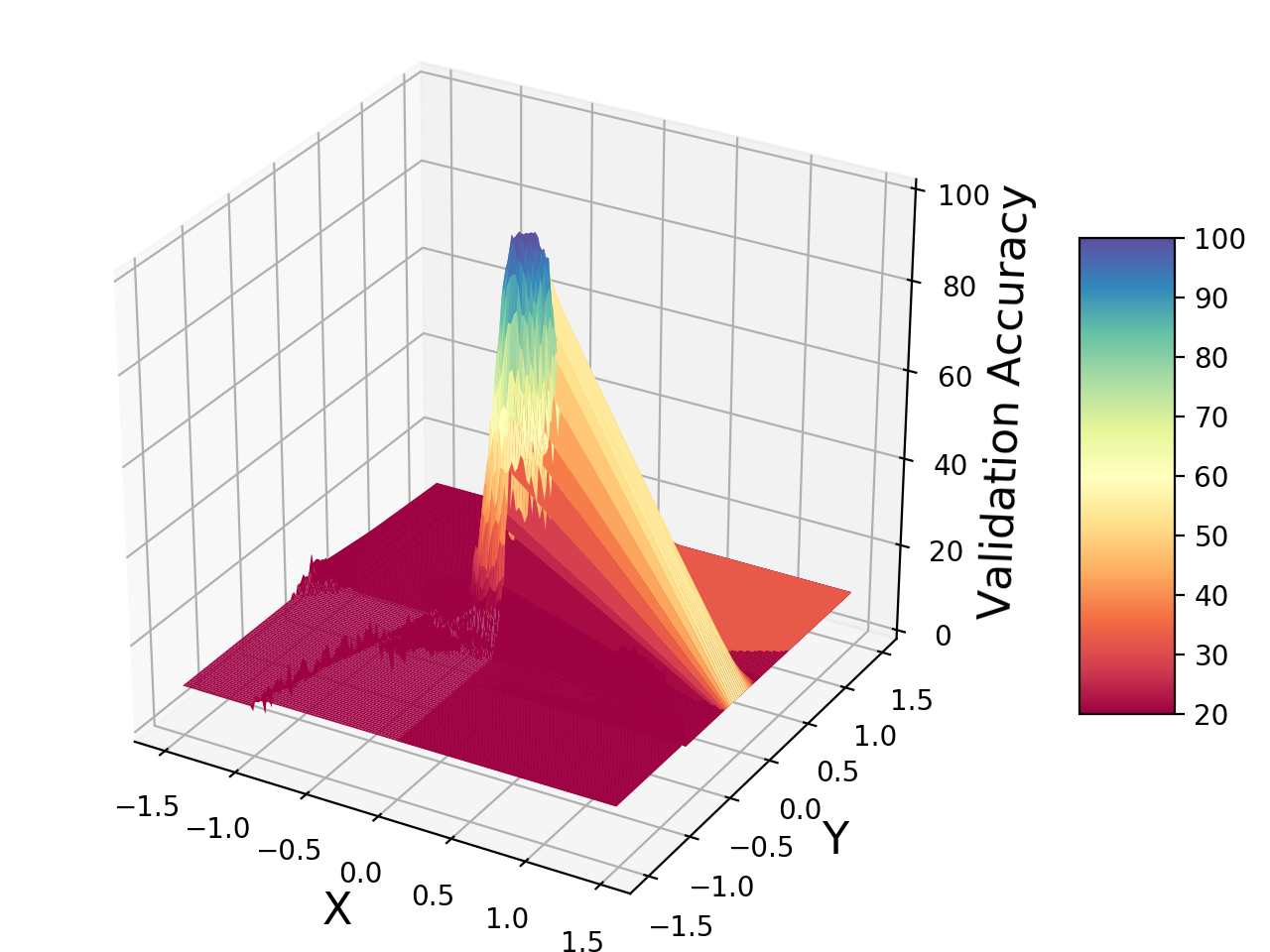}}
  \subfigure[\EXP]{\includegraphics[width=0.33\linewidth]{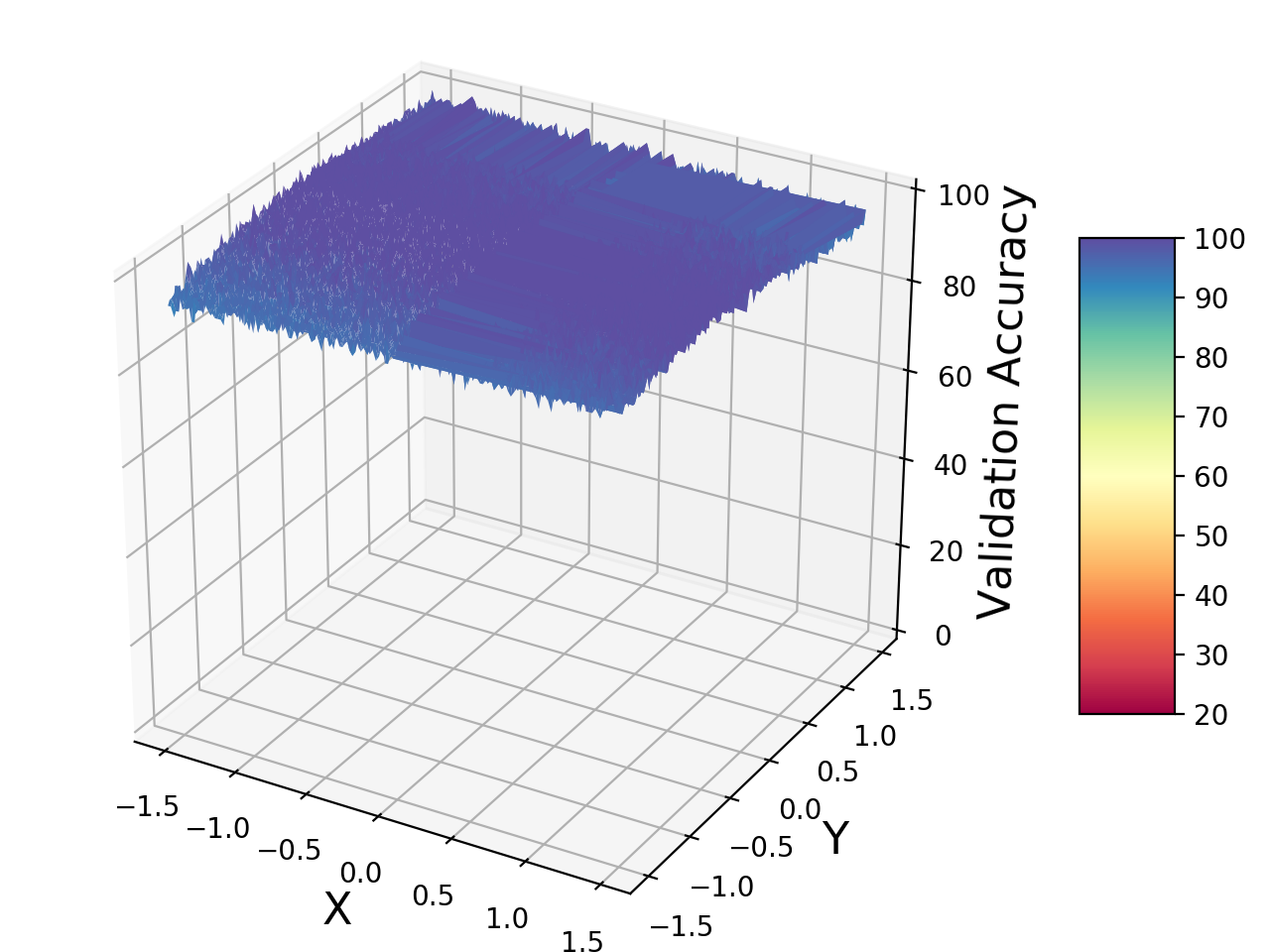}}
  \subfigure[\ADV]{\includegraphics[width=0.33\linewidth]{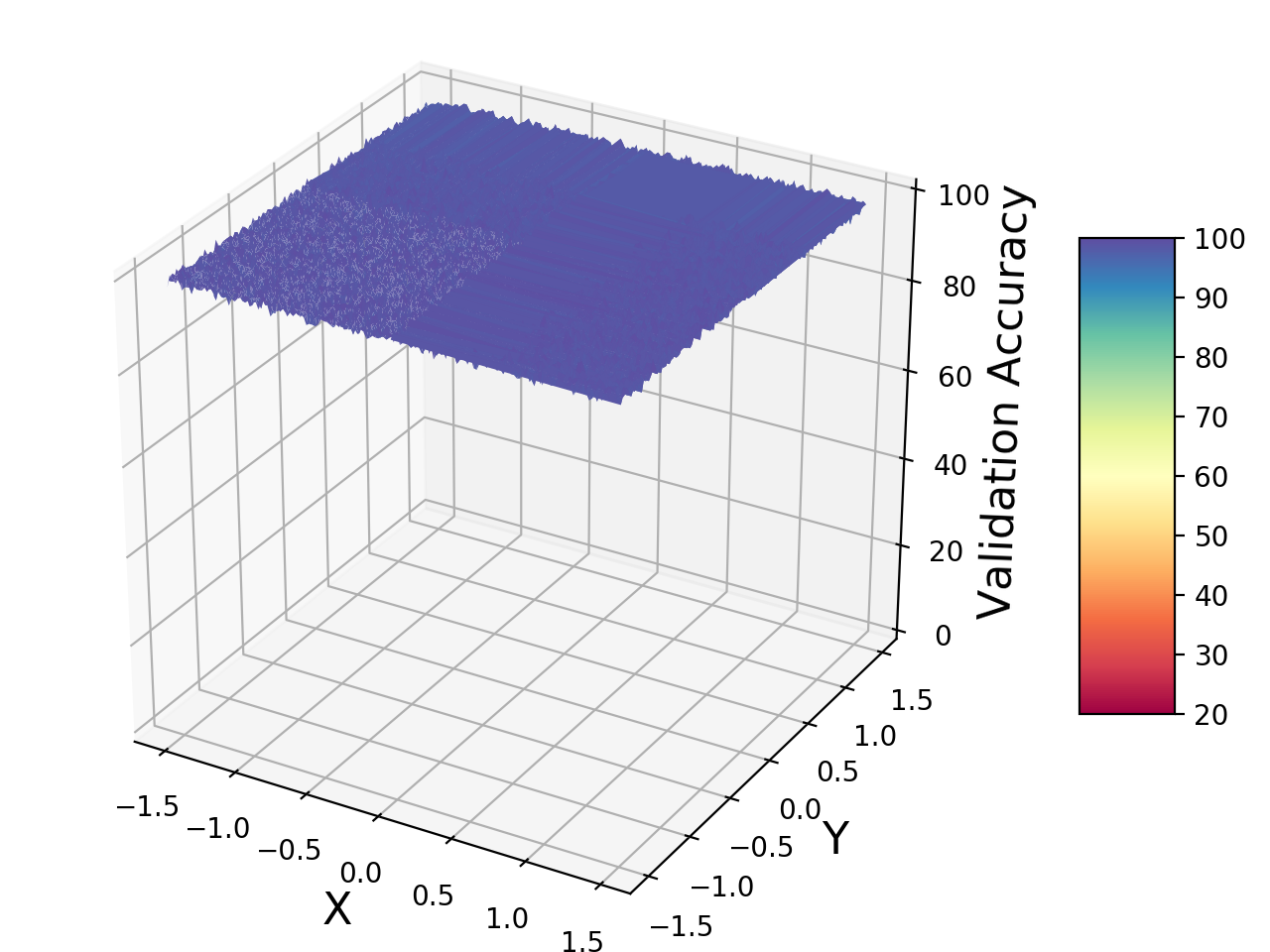}}
  \vskip 0.1in
  \caption{The landscape of validation accuracy w.r.t. the architectural parameters $\alpha$ of DARTS \cite{liu2018darts}, SDARTS-RS and SDARTS-ADV. Figure from Chen et al. \cite{chen2020stabilizing}.}
  \label{fig:sdarts}
\end{figure}

\textbf{BigNAS} \cite{yu2020bignas} is an original approach where the authors proposed to palliate the optimization gap by directly reusing the weights of the supernet/one-shot model to evaluate the performance of the final architecture. This contrasts with previous works \cite{liu2018darts, cai2018proxylessnas, cai2020once} that either retrained the network weights or preprocessed them in some way. To achieve this, BigNAS performs single-stage training using adapted versions of existing training methods such as inplace distillation \cite{yu2019universally}, the sandwich rule \cite{yu2019universally}, exponential learning rate decay scheduling, or dropout-based regularization \cite{tan2021efficientnetv2}. These modifications aim to stabilize training and enable BigNAS to efficiently train both large and small candidate architecture within its supernet. Furthermore, the authors proposed a coarse-to-fine architectural selection scheme where a skeleton architecture is first selected according to specific sets of requirements (e.g., input resolution, network depth, or kernel size). Then, these sets are fine-tuned with random mutations to obtain an optimal architecture. BigNAS reaches up to 80.9 \% top-1 accuracy on ImageNet \cite{deng2009imagenet} for its largest model (9.5 M parameters, 1 GFLOPS), thus overperforming previous approaches. 

Vahdat et al. \cite{Vahdat_2020_CVPR} proposed to combine DNAS and Reinforcement Learning-based NAS in a unified framework, denoted \textbf{UNAS}, that would bring out the strengths of both approaches. This way, UNAS can search for both differentiable and non-differentiable objectives. In particular, they combine a corrected variant of the classical REINFORCE RL algorithm \cite{williams1992simple} with a Gumbel-Softmax \cite{jang2017categorical} sampled DNAS algorithm to jointly search for either a differentiable or a non-differentiable objective. Hence, the gradient of a differentiable loss $\mathcal{L}_d$ can be computed as
\begin{equation}
\begin{split}
    \Delta_{\alpha}\mathcal{L}_d & = \text{REINFORCE}(\mathcal{L}_d, c_d(\alpha)) + C( c_d(\alpha)) + \text{gumbel\_softmax}(\alpha, c_d(\alpha))\\
    & = \mathbb{E_{p_\phi(\alpha)}}\left[(\mathcal{L}_d(\alpha)-c_d(\alpha))\frac{\partial\text{log}p_{\phi}(\alpha)}{\partial \phi}\right] - \mathbb{E_{p_\phi(\alpha)}}\left[\frac{\partial c_d(\alpha)}{\partial \phi}\right] + \frac{\partial \mathbb{E_{p_\phi(\alpha)}}\left[c_d(\alpha)\right]}{\partial \phi},
\end{split}
\end{equation}
where $c_d$ is a control variate defined as $c_d(\alpha) = \mathbb{E_{r_\phi(\zeta|\alpha)}}\left[\mathcal{L}_d(\zeta)\right]$ used to lower the high variance of REINFORCE where $\zeta$ is a smooth architecture sampled from a conditional Gumbel-Softmax distribution $r_\phi(\zeta|\alpha)$.
In addition, UNAS can help bridge the optimization gap by introducing a novel objective function $\mathcal{L}_{gen}$ to avoid architectural overfitting by taking into account the gap between $\mathcal{L}_{train}$ and $\mathcal{L}_{val}$ in the optimization process. $\mathcal{L}_{gen}$ is defined as follows: 
\begin{equation}
    \mathcal{L}_{gen}(\alpha, w) = \mathcal{L}_{train}(\alpha,w) + \lambda|\mathcal{L}_{val}(\alpha, w)-\mathcal{L}_{train}|
\end{equation}
where $\alpha$ denotes the architectural parameters, $w$ represents the network weights, and $\lambda$ is a coefficient weighting the generalization gap. UNAS overperforms previous DNAS works on CIFAR-10/100 and ImageNet while maintaining a search cost comparable to DARTS (~4 GPU days).

In addition to the optimization gap issue, Zhang et al. \cite{zhang2020differentiable} observed that a catastrophic forgetting problem (multi-model forgetting \cite{benyahia2019overcoming})  occurs in the supernet's weights training, leading to a deterioration of the optimization process for all the candidate architectures derived from the supernet. To palliate these issues, they introduced $\textbf{E}^2\textbf{NAS}$ (Exploration Enhancing Neural Architecture Search with Architecture Complementation), a novel DNAS approach that leverages a VGAE (Variational Graph AutoEncoder) to create an injection between the final discrete architectures and the continuous search space. More precisely, an asynchronous message-passing scheme encodes the architecture into an injective space by encoding the final output $C$ of the network into a continuous representation $z$ (i.e., a latent space). Hence, the hidden state $h_v$ of node $v$ is defined as
\begin{equation}
    h_v = \mathcal{U}(w_v, h^{in}_{v})
    \text{ with } h^{in}_{v} = \mathcal{G}({h_u: u \rightarrow v}),
\end{equation}
where $\mathcal{U}$ is a function updating the hidden state $h^{in}_{v}$ obtained by aggregating all its predecessors with function $\mathcal{G}$. Since both $\mathcal{G}$ and $\mathcal{U}$ are injective, the VGAE maps $C$ to $z$ injectively. From then, $\text{E}^2\text{NAS}$ performs differentiable architecture search on the latent continuous space. In addition, a new complementation loss $\mathcal{L}_c$ is introduced to tackle the catastrophic forgetting problem. This loss works in conjunction with a replay buffer that contains the last architecture $\alpha_{i-1}$ along with another complementary architecture $\alpha_{i}^c$. $\mathcal{L}_{train}$ in Eq. \ref{eq:bilevel_optimization} is replaced with $L_c$ defined for weights $w_i = w^*(\alpha_i)$ and $w_{i}^c = w^*(\alpha_{i}^c)$ at step $i$ as
\begin{equation}
    \mathcal{L}_c(w_i) = (1-\epsilon)\mathcal{L}_{CE} + \epsilon(\mathcal{L}_{CE}(w_{i}^c) + \mathcal{L}_{CE}(w_{i-1})) + \eta\mathcal{R}(w_i),
\end{equation}
where $\mathcal{L}_{CE}$ is the Cross-Entropy loss, $\mathcal{R}$ is a $l_2$ regularization term, $\epsilon$ is a value that balances between optimizing the current architecture (exploitation) or preventing other alternatives from vanishing (exploration). $\text{E}^2\text{NAS}$ successfully overperformed previous works on the three datasets available in NAS-Bench-201 \cite{dong2020nasbench201} (e.g., a +29.45 \% top 1 accuracy improvement on ImageNet-16-120).

Zhang et al. \cite{zhang2021idarts} proposed \textbf{iDARTS}, a solution that reformulates the optimization process of DARTS with a Neumann-approximation of the Implicit Theorem Function (IFT) \cite{lorraine2020optimizing}. Concretely, the architectural parameter gradients $\Delta_{\alpha}\mathcal{L}_{val}$ (see Eq. \ref{eq:gradient_calc}) are calculated as follows:
\begin{equation}
    \Delta_{\alpha}\mathcal{L}_{val} =  \frac{\partial \mathcal{L}_{val}}{\partial \alpha}-\frac{\partial \mathcal{L}_{val}}{\partial w}\left[\frac{\partial^{2} \mathcal{L}_{train}}{\partial w^2}\right]^{-1}\frac{\partial^{2}\mathcal{L}_{train}}{\partial \alpha \partial w},
    \label{eq:ift}
\end{equation}
where $\mathcal{L}_{val}$ is the validation loss, $\mathcal{L}_{train}$ is the training loss, and $w$ are the network weights.
However, it is computationally intensive to compute the inverse of the Hessian matrix $\frac{\partial^{2} \mathcal{L}_{train}}{\partial w \partial w}$ in Eq. \ref{eq:ift}. Hence, to alleviate this burden, the authors approximated this inverse matrix using a Neumann series \cite{lorraine2020optimizing}. This Neumann approximation is computed in a stochastic setting where minibatches are used instead of the whole dataset. Thus, the stochastic approximation of the gradients described in Eq. \ref{eq:ift} is formulated as follows:
\begin{equation}
    \Delta_{\alpha}\hat{\mathcal{L}}^{i}_{val}(w^{j}(\alpha), \alpha) =  \frac{\partial \mathcal{L}^{i}_{val}}{\partial \alpha}-\gamma \frac{\partial \mathcal{L}^{i}_{val}}{\partial w}\sum_{k=0}^{K}\left[I-\frac{\partial^{2} \mathcal{L}^{j}_{train}}{\partial w^2}\right]^{k}\frac{\partial^{2}\mathcal{L}^{j}_{train}}{\partial \alpha \partial w},
\end{equation}
where $i$ and $j$ are minibatches randomly sampled from the training and validation datasets respectively, $\gamma$ is the learning rate, $K$ is the number of terms of the Neumann series used for approximation, and $I$ is the identity matrix. This reformulation of the architectural gradient computation performs multiple optimization steps before updating $\alpha$, hence making $w(\alpha)$ closer to its optimal value $w^*(\alpha)$. The authors empirically showed that iDARTS improves performance over standard DARTS by 2.6 \% on ImageNet \cite{deng2009imagenet}.

\begin{figure}[htbp]
  \centering
  \subfigure{\includegraphics[clip, width=0.33\textwidth]{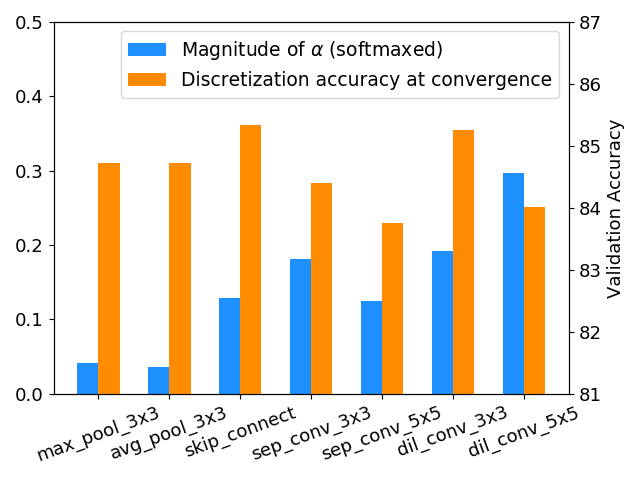}}
  \subfigure{\includegraphics[clip, width=0.33\textwidth]{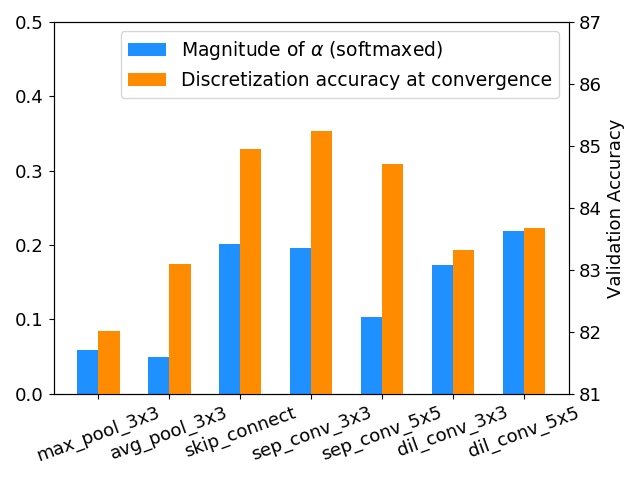}}
  \subfigure{\includegraphics[clip, width=0.33\textwidth]{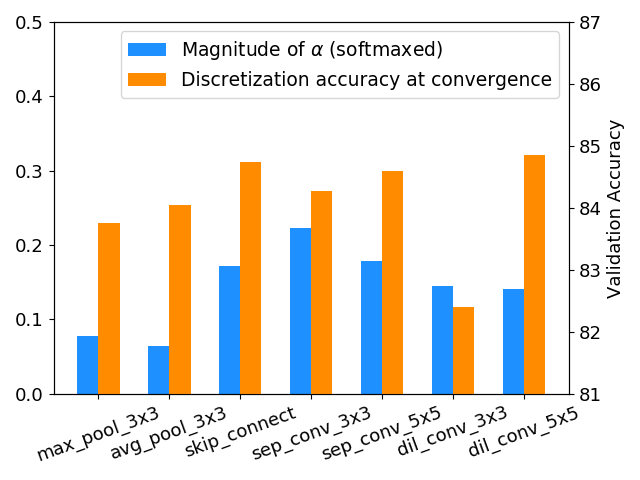}}
  \caption{Bar chart comparing the value of $\alpha$ w.r.t. the discretization accuracy at convergence for each operation of 3 randomly selected edges from a pretrained DARTS model. Figure from Wang et al. \cite{wang2021rethinking}.}
  \label{fig:dartspt}
\end{figure}

Wang et al. \cite{wang2021rethinking} argued that the optimization gap in DARTS is linked to the architecture selection process, as the $\alpha$ weight values associated with an operation might not always reflect this operation's strength. They defined the discretization accuracy at convergence of an operation as the supernet accuracy after discretizing to this operation and fine-tuning the remaining network until it converges again. Hence, Fig. \ref{fig:dartspt} showcases empirical evidence that the discretization accuracy at convergence of an operation does not necessarily match its $\alpha$ weight value. In fact, some operations with a small $\alpha$ value can reach a high discretization accuracy, further reinforcing the critical aspect of the underlying architecture selection problem. 

To alleviate this issue, the authors proposed a perturbation-based architecture selection (PT) where each operation of each edge of the architecture is masked in turn. Then, the operation that leads to the highest drop in performance when masked is considered to be the most important on that edge. This process is not too invasive as it only masks one operation at a time, thus making the supernet accuracy close to the one of the unmodified supernet. Finally, the authors showed that training a DARTS supernet normally and then using PT to discretize the architecture (a process denoted \textbf{DARTS+PT}) significantly improves performance (e.g., +0.4 \% top 1 accuracy on CIFAR-10 compared to DARTS). 

\begin{figure}[h]
  \centering
  \includegraphics[width=0.7\linewidth]{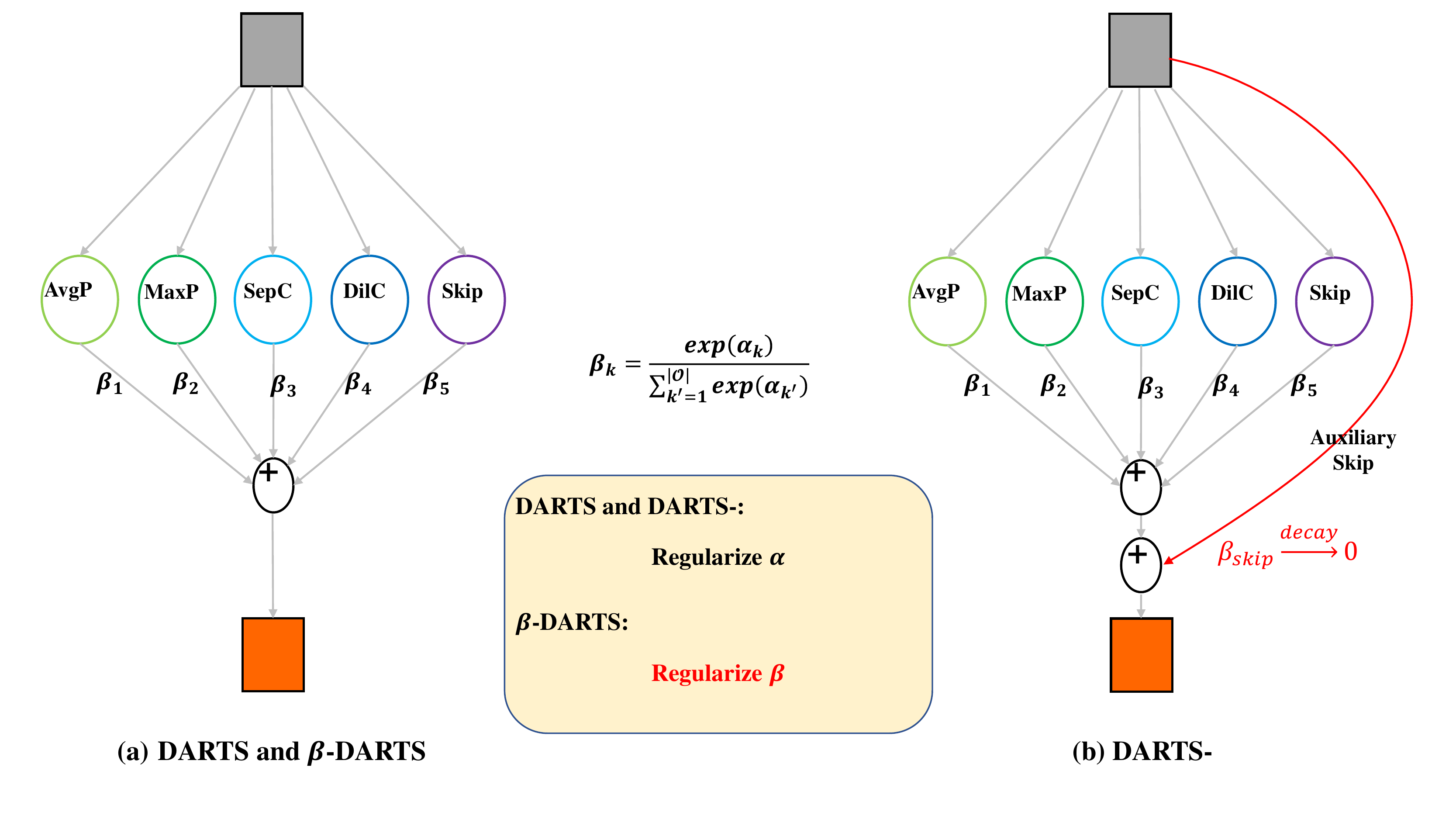}
  \caption{Layout of the $\beta$-DARTS search process in comparison with DARTS \cite{liu2018darts} and DARTS- \cite{chu2020darts}. Figure from Ye et al. \cite{ye2022b}}
  \label{fig:beta_darts}
\end{figure}

In the continuation of DARTS- (discussed in Section \ref{sec:skip_connections}), \textbf{$\beta$-DARTS} \cite{ye2022b} introduced a novel and very simple regularization method called Beta-Decay inspired from $\mathcal{L}_2$ regularization that involves imposition restrictions on the architectural parameters to reduce optimization discrepancies. This regularization occurs on the $\alpha$ parameters after the \textit{softmax} activation and consists of a straightforward loss function $\mathcal{L}_{Beta}$:
\begin{equation}
    \mathcal{L}_{Beta}(\alpha) = log (\sum_{k=1}^{K}\text{exp}(\alpha^{k})),
\end{equation}
where $K$ is the total number of candidate operations. $\mathcal{L}_{Beta}$ is differentiable and added to the validation loss $\mathcal{L}_{val}$ pondered by a parameter denoted $\lambda$. Thus, Eq. \ref{eq:bilevel_optimization} is modified as follows:
\begin{equation}
    \underset{\alpha}{min}(\mathcal{L}_{val}(w^*(\alpha), \alpha) + \lambda \mathcal{L}_{Beta}(\alpha)).
\end{equation}
According to the theoretical analysis provided by the authors, $\mathcal{L}_{Beta}$ improves generalization and increases robustness. Ultimately, $\beta$-DARTS reached competitive scores on both small-scale (CIFAR-10/100) and large-scale (ImageNet) datasets while searching only on CIFAR-10 and CIFAR-100. The search process of $\beta$-DARTS is resumed in Fig. \ref{fig:beta_darts}.

\begin{figure}[h]
  \centering
  \begin{overpic}[width=\linewidth]{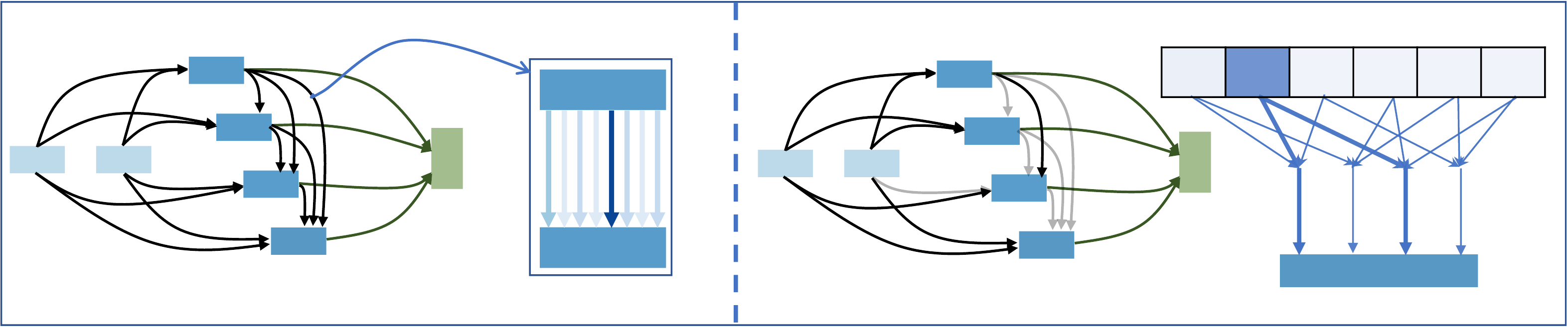}
			\put(1.7, 10.25){$\mathbf{x_1}$}
			\put(6.9, 10.25){$\mathbf{x_2}$}		
			\put(12.9, 16){$\mathbf{x_3}$}
			\put(14.6, 12.4){$\mathbf{x_4}$}
			\put(16.3, 8.8){$\mathbf{x_5}$}
			\put(18.1, 5.1){$\mathbf{x_6}$}
			\put(37.5, 14.7){$\mathbf{x_3}$}
			\put(37.5, 4.5){$\mathbf{x_6}$}		
			\put(27.8, 11.6){\rotatebox{-90}{$\mathbf{x_7}$}}
			
			\put(49.1, 10){$\mathbf{x_1}$}
			\put(54.3, 10){$\mathbf{x_2}$}		
			\put(60.3, 15.75){$\mathbf{x_3}$}
			\put(62, 12.15){$\mathbf{x_4}$}
			\put(63.7, 8.55){$\mathbf{x_5}$}
			\put(65.5, 4.95){$\mathbf{x_6}$}
			\put(75.5, 11.3){\rotatebox{-90}{$\mathbf{x_7}$}}
			\put(87,3.2){$\mathbf{x_5}$}
			\footnotesize
			\put(53.5, 8){$\mathbf{1}$}
			\put(59, 8.8){$\mathbf{2}$}		
			\put(64.4, 10.2){$\mathbf{3}$}
			\put(64.9, 13.4){$\mathbf{4}$}
			\normalsize
			\put(81.6, 7){$\mathbf{1}$}
			\put(85, 7){$\mathbf{2}$}		
			\put(88.3, 7){$\mathbf{3}$}
			\put(91.8, 7){$\mathbf{4}$}
			
			\put(75, 15.5){$\mathbf{12}$}
			\put(79, 15.5){$\mathbf{13}$}
			\put(83, 15.5){$\mathbf{14}$}
			\put(87, 15.5){$\mathbf{23}$}
			\put(91, 15.5){$\mathbf{24}$}
			\put(95, 15.5){$\mathbf{34}$}
			\large
			\put(1, 18.5){\textbf{Operation Search}}
			\put(48, 18.5){\textbf{Topology Search}}
  \end{overpic}
  \vspace{-.3in}
  \caption{Layout of the DOTS search process featuring both the operational and topological subprocesses. Figure from Gu et al. \cite{gu2021dots}}
  \label{fig:dots}
\end{figure}
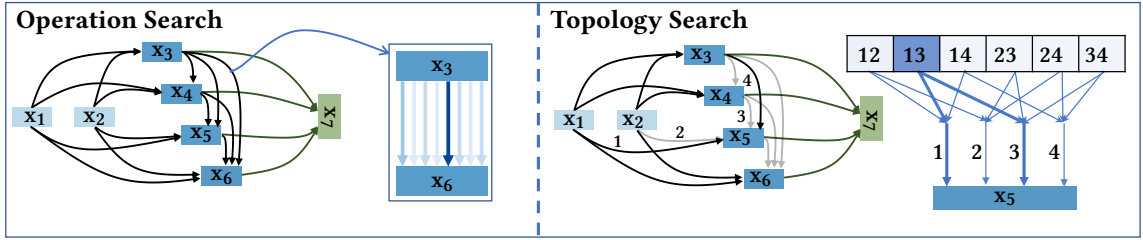

Gu et al. \cite{gu2021dots} argued that the ranking of operations in DARTS edges is not representative of the final model performance as it does not take correctly into account operations that are related to topology (e.g., \textit{skip connections}), hence there is an optimization gap between the proxy model and the final model. To solve this issue, they proposed the novel concept of decoupling the operation and topology search that are performed simultaneously in the original DARTS. This solution, named \textbf{DOTS}, divides the search process into two stages. First, during the topology search stage, the topology search space $\mathcal{E}$ is continuously relaxed into topology weights that are associated with pairwise combinations of edges. For instance, considering node $x_j$, $\mathcal{E}_{x_j}$ is defined as follows:
\begin{equation}
    \mathcal{E}_{x_j} = \{\langle e_{i_{1},j}, e_{i_{2},j}\rangle|0 < i_1 < i_2 < j\}.
\end{equation}
Moreover, for each edge $e_{i,j}$, weights of combinations containing this edge are aggregated into $\gamma_{i,j}$ weights to reduce the search cost. $\gamma_{i,j}$ is defined by the following equation:
\begin{equation}
    \gamma_{i,j} = \sum_{c \in \mathcal{E}_{i,j}, e_{i,j} \in c} \frac{exp(\beta^{c}_{x_j}/T_\beta)}{N(c)\sum_{c' \in \mathcal{E}_{x_j}} exp(\beta^{c'}_{x_j}/T_\beta)},
\label{eq:dots}
\end{equation}
where $N(c)$ is the number of edges in edge combination $c$, and $\beta^{c}_{x_j}$ represents the weight associated with $c$. Eq. \ref{eq:dots} uses an architectural annealing scheme with temperature $T_{\beta}$ as previous works \cite{noy2020asap, xie2018snas} found that this mechanism helps to bridge the optimization gap when searching. In the second phase, DOTS performs an operation search to select the single optimal operation for each edge according to architectural weights $\alpha$ (similarly to DARTS). However, this strategy could drop some topology-oriented operations before the topology search, thus altering the optimization process. To prevent this, DOTS introduced a group strategy where the operation search space $O$ is divided into $p$ subspaces on which the search process is performed independently. Once the search process is over, the best operation from each subspace is selected and merged into a new operation search space. The authors showed that this group strategy effectively preserves topology-related and topology-agnostic operations. DOTS successfully overperformed DARTS top 1 accuracy scores by +0.63 \% on CIFAR-10 and by +2.7 \% on ImageNet. The global process of DOTS is summarized in Fig. \ref{fig:dots}.

Yang et al. \cite{yang2021towards} introduced \textbf{EnTranNAS} as a different solution to the optimization gap problem. EnTranNAS comprises Engine-cells (standard DARTS-like differentiable cells) and Transit-cells (transits the derived/discretized architecture). It only searches for a single cell, as the author argues it is sufficient to perform DNAS. Contrary to DARTS, the architecture discretization process in EnTranNAS is no longer part of post-processing but rather done at the end of each search iteration. Hence, the Transit-cells serve to host the currently derived architecture and transmit it to later cells. EnTranNAS includes the target (derived) architecture in the search process, resulting in higher confidence when selecting operations. In addition, the authors introduced a feature-sharing strategy to improve search efficiency, assuming that the same operation from node $i$ to node $j> i$ always shares the same features in a single cell. Thus, Eq. \ref{eq:mixed_output} is modified as follows
\begin{equation}
    \overline{o}_{i,j}(x) = \begin{cases}
        \sum_{i<j} \sum^{K}_{k=1} \frac{exp(\alpha^{k}_{i,j}/\tau)}{\sum^{K}_{k'=1} exp(\alpha^{k'}_{i,j}/\tau)}o^{k}_{i,j}(x) = p^{k}_{i,j}o^{k}_{i,j}(x), \text{ in Engine-cells,}\\
        \sum_{(i,k)\in S_j} o^{k}_{i,j}(x), \text{ in Transit-cells,}\\
    \end{cases}
    \label{eq:entrannas}
\end{equation}
where $\tau$ is a temperature parameter that acts as a regularization factor for the differentiable process in the Engine-cells. This strategy helps to balance optimization between parametric and non-parametric operations. It reduces the computational cost by only computing feature maps of each operation only once per cell (from node $i$ to ulterior nodes $j>i$). However, EnTranNAS does not completely eliminate the optimization gap. Hence the authors also proposed a novel topology-search-oriented architecture derivation method dubbed EnTranNAS-DST. Concretely, they introduced an additional set of trainable parameters $\{\beta_{j}\}_{j=2}^n$ for each intermediary node $j$ and implemented thresholds $t_{j} = \text{sigmoid}(\beta_j)$ to perform operation pruning on those nodes as
\begin{equation}
    q^{k}_{i,j} = \text{ReLU}(\frac{p^{k}_{i,j}}{max_{i<j,1\leq k \leq K}\{p^{k}_{i,j}\}}-t_{j}).
\end{equation}
If there is $k$ s.t. $q^{k}_{i,j} \neq 0$, $q^{k}_{i,j}$ is further normalized by
\begin{equation}
    \hat{q}^{k}_{i,j} = \frac{q^{k}_{i,j}}{\sum_{k} q^{k}_{i,j}}.
\end{equation}
EnTranNAS-DST node output is thus obtained simply by replacing $p^{k}_{i,j}$ with $\hat{q}^{k}_{i,j}$ in Eq. \ref{eq:entrannas}. The authors experimentally showed that EnTranNAS overperforms most prior works on both CIFAR-10 (+0.28 \% top 1 accuracy vs. DARTS) and ImageNet (+2.9 \% top 1 accuracy compared to DARTS).

\begin{figure}[h]
  \centering
  \includegraphics[width=0.5\linewidth]{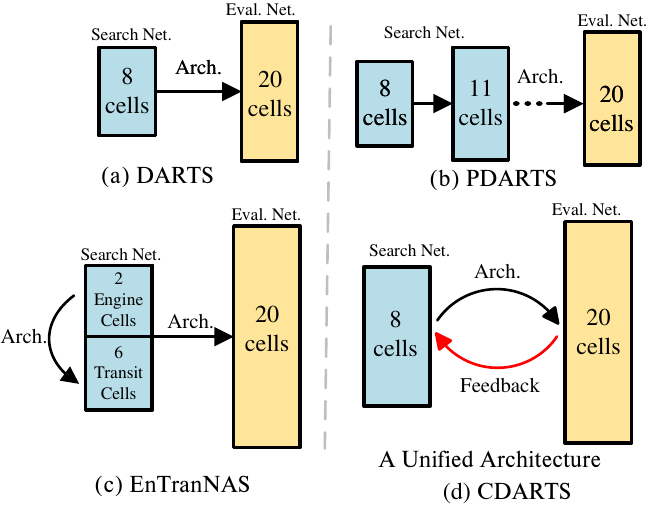}
  \caption{Comparison between the search processes of DARTS \cite{liu2018darts}, P-DARTS \cite{chen2019progressive}, EnTranNAS \cite{yang2021towards}, and CDARTS \cite{yu2022cyclic}. Figure from Yu et al. \cite{yu2022cyclic}}
  \label{fig:cdarts}
\end{figure}

\textbf{CDARTS} \cite{yu2022cyclic} proposed to address the optimization gap issue by implementing a cyclic feedback mechanism between the search and evaluation networks analogous to a teacher-student model. The search network (composed of 8 cells) provides an intermediate architecture to the evaluation network (composed of 20 cells) and, in return, gets performance feedback. Hence, the search strategy takes into account the performance of the final discretized (and larger) architecture. Furthermore, the two networks are jointly trained and unified into a single architecture. The joint optimization problem is defined as:
\begin{equation}
\underset{\alpha}{\text{min }} \mathcal{L}_{val}(w_{E}^{*}, w_{S}^{*}, \alpha) \quad \text{s.t. } \begin{cases}
    w_{E}^{*} = \underset{w_E}{\text{argmin }} \mathcal{L}_{val}(w_{E},\alpha),\\
    w_{S}^{*} = \underset{w_S}{\text{argmin }} \mathcal{L}_{train}(w_{S},\alpha),\\
\end{cases}
\end{equation}
where $w_S$ and $w_E$ are the weights of the search and evaluation networks respectively. CDARTS' search process comprises two stages. Firstly, the separate learning stage during which both networks are trained individually on the input dataset. $\alpha$ weights are initialized with random values, whereas the cell architectures of the evaluated are initialized from the top-$k$ discretization $\overline{\alpha}$ of the learned $\alpha$. Secondly, the joint optimization stage where the search algorithm leverages performance feedback from the evaluation network to update $\alpha$ defined as follows:
\begin{align}
    \alpha^*, w_{E}^* = \underset{\alpha,w_E}{\text{argmin }} \mathcal{L}_{val}^S(w_{S}^*, \alpha)+\mathcal{L}_{val}^E(w_E,\overline{\alpha}) + \lambda\mathcal{L}_{val}^{S,E}(w_{S}^*,\alpha,w_E,\overline{\alpha}),
\end{align}
where $\mathcal{L}_{val}^{S,E}$ denotes the knowledge transfer procedure between the search and evaluation networks, dubbed \textit{introspective distillation} and formulated as:
\begin{equation}
    \mathcal{L}_{val}^{S,E} (w_{S}^*,\alpha,w_{E},\overline{\alpha}) = \frac{T^2}{N}\sum_{i=1}^N p(w_E,\overline{\alpha})\text{log}(\frac{p(w_E,\overline{\alpha})}{q(w_{S}^*, \alpha)})
\end{equation}
where $N$ is the number of training samples, $T$ is a temperature coefficient, and $p$ and $q$ are the output feature logits of the evaluation and search networks respectively (computed using a \textit{softmax}). CDARTS overperforms previous methods on DARTS search space (e.g., + 3\% top 1 accuracy improvement compared to DARTS) while keeping the computational cost reasonable (1.7 GPU days). The main concept behind CDARTS is showcased in Fig. \ref{fig:cdarts}.

\subsection{Over-representation of \textit{skip connections} in DARTS}
\label{sec:skip_connections}
\begin{figure}[h]
  \centering
  \includegraphics[scale=0.5]{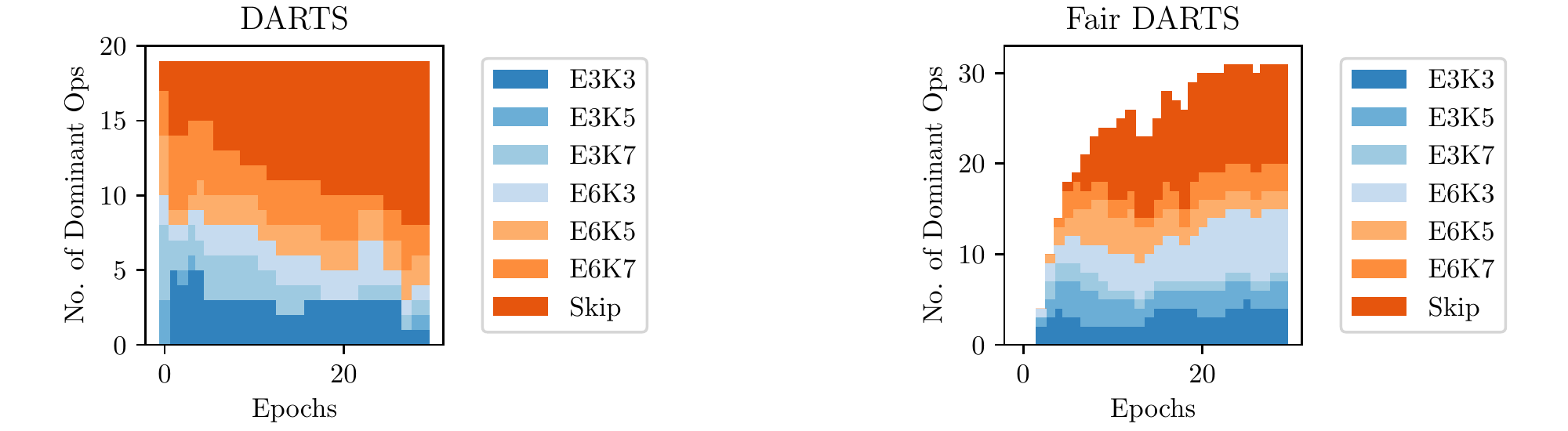}
  \includegraphics[width=0.45\textwidth,scale=0.6]{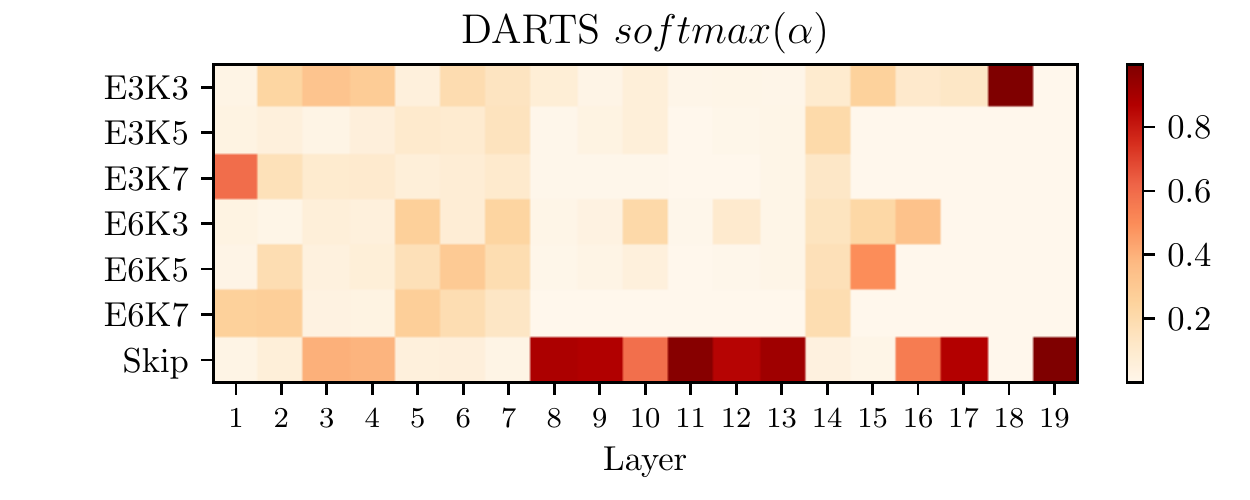}
  \includegraphics[width=0.45\textwidth,scale=0.6]{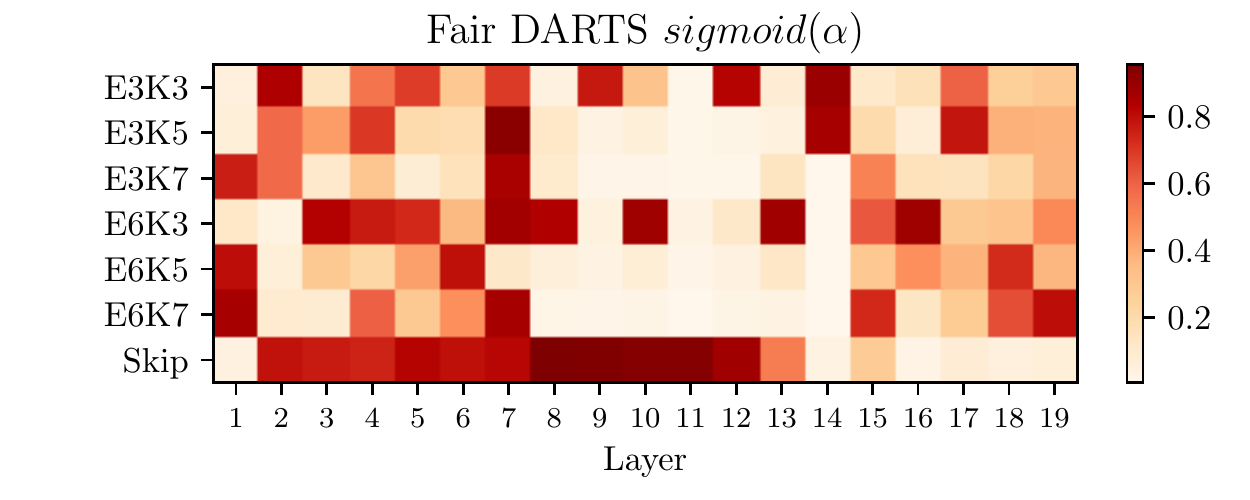}
  \caption{Stacked area plot of the number of dominant operations of DARTS and FairDARTS when searching on ImageNet. Figure from Chu et al. \cite{chu2020fair}}
  \label{fig:fairdarts}
\end{figure}

As already discussed in Section \ref{sec:gradient_approximation}, \textbf{FairDARTS} \cite{chu2020fair} replaced \textit{softmax} with the \textit{sigmoid} operation $\sigma$ to ensure fair competition between the different operations (i.e., the weights associated with operations can independently increase or decrease). This means that a high prominence of \textit{skip connections} will not suppress the other operations that can thus overperform and replace them. Empirically, this results in a lessened presence of \textit{skip connections} in the final architectures as shown in Fig. \ref{fig:fairdarts}. Some other methods, such as D-DARTS \cite{heuillet2021d} (detailed in Sec. \ref{sec:search_space}), tackled the \textit{skip connections} issue using the same approach as FairDARTS.

In a different manner, the authors of \textbf{P-DARTS} \cite{chen2019progressive} managed to restrict the number of \textit{skip connections} by introducing an operation-level \textit{dropout} \cite{srivastava2014dropout} to regularize the search space. More accurately, the \textit{dropout} mechanism is placed after every \textit{skip connection} to block the path and entice the search algorithm to explore other operations. In addition, the \textit{dropout} rate is gradually decayed to prevent the \textit{skip connections} from being completely suppressed (i.e., \textit{skip connections} are heavily penalized at the start of the search process and then treated equally with the other operations at the end). Additional details on P-DARTS can be found in Section \ref{sec:gradient_approximation}.

\begin{figure}[h]
  \centering
  \includegraphics[width=0.6\linewidth]{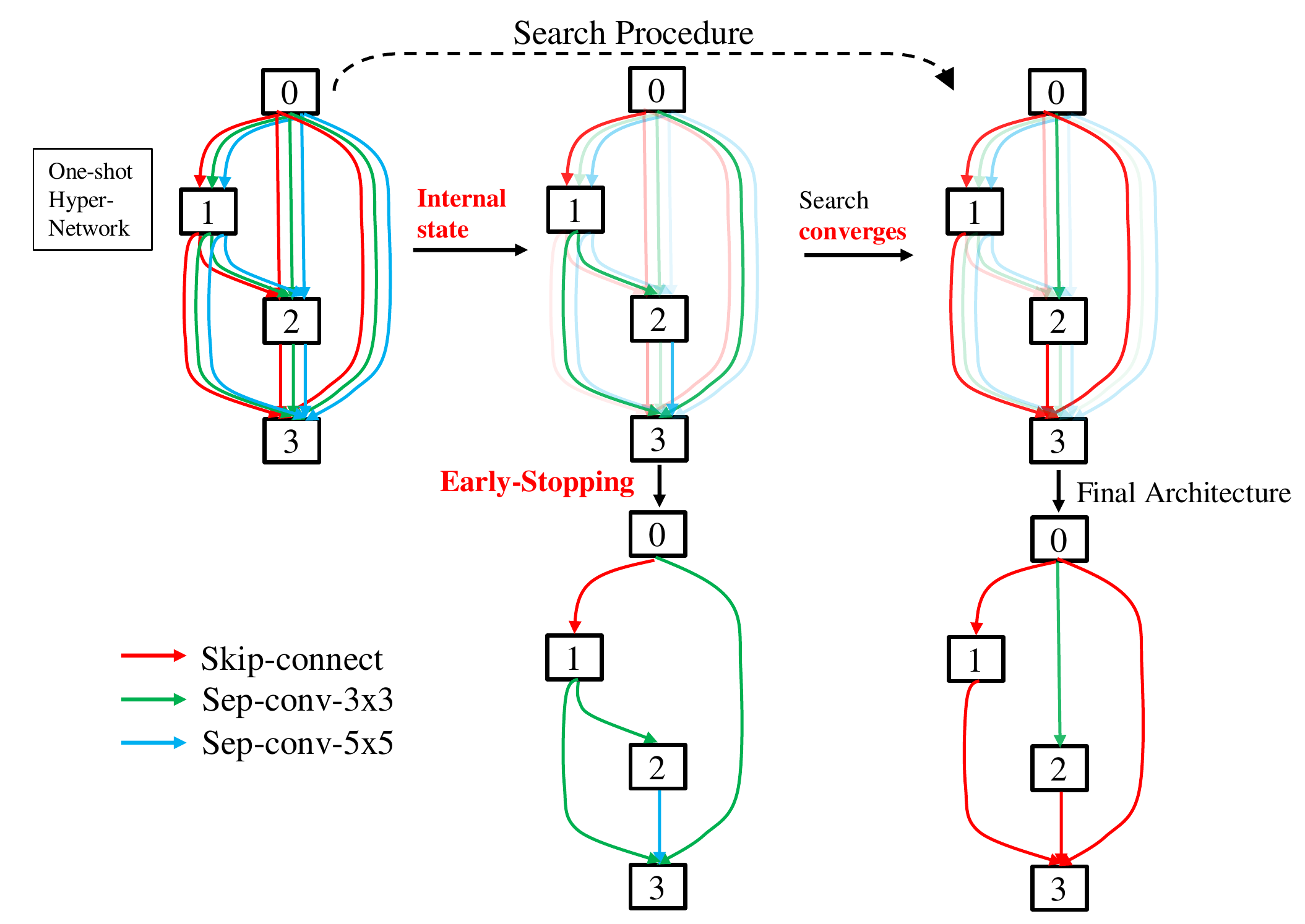}
  \caption{Illustration of the early stopping process in DARTS+. Figure from Liang et al. \cite{liang2019darts+}}
  \label{fig:darts+}
\end{figure}

Liang et al. \cite{liang2019darts+} argued that the over-representation of \textit{skip connections} results from an overfitting phenomenon in the optimization process of DARTS. To alleviate this issue, they proposed \textbf{DARTS+}, an early-stopping procedure that ends the search phase if the following criteria are met:
\begin{enumerate}
    \item \textit{Two or more skip connections are present in the normal cell architecture.}
    \item \textit{The ranking of architecture parameters $\alpha$ for learnable operations becomes stable for a determined number of epochs (e.g., 10 epochs).}
\end{enumerate}
The authors showed that using either of these criteria led to performance improvements over previous baselines (e.g., + 0.7 \% top 1 accuracy compared to DARTS when using Criterion 1). Furthermore, they provided empirical evidence that Criterion 1 is easier to use and implement but yields less accurate results than Criterion 2. This simple early stopping procedure was dubbed DARTS+ and is illustrated by Fig. \ref{fig:darts+}.

Zela et al. \cite{Zela2020Understanding} also focused on robustifying DARTS as they found out that performance collapses in many cases with high dominance of unparameterized (i.e., \textit{skip}/\textit{pooling}) operations. Hence, they proposed DARTS-ES, a novel method that performs early stopping according to the eigenvalues of the Hessian matrix of the validation loss $\Delta^{2}_{\alpha} \mathcal{L}_{val}$ w.r.t. the $\alpha$ weights. More precisely, they showed that large eigenvalues often lead to degenerate architectures and tracked these values to stop the search process before the performance collapses. Furthermore, they implemented two different regularization methods. The first one uses a combination of well-known data augmentations techniques (Cutout \cite{devries2017improved}, and ScheduledDropPath \cite{zoph2018learning}). The second one increases $\mathcal{L}_2$ regularization by choosing among several factors (e.g., ${1,3,9,27,81}$). Both techniques successfully increased robustness, especially when combined with DARTS-ES, an approach dubbed \textbf{R-DARTS}. The authors tested their approach on several computer vision datasets (CIFAR10/100 \cite{krizhevsky2009learning}, SVHN \cite{netzer2011reading}) and under different search spaces. R-DARTS improved top 1 accuracy on CIFAR-10 up to +3.64 \% compared to DARTS.

As a non-DARTS approach, $\textbf{E}^2\textbf{NAS}$ \cite{zhang2020differentiable} (first presented in Section \ref{sec:gradient_approximation}) also addressed the over-representation of non-parametric operations in their own way. They tackled \textit{the rich-get-richer problem}, in which the optimizer is biased towards architectures with high performance in their early stage. They added a measure of the novelty into the gradient to avoid being stuck in local minima, hence computing architectural weights $\alpha_{\theta}$ update as
\begin{equation}
    \alpha^{i+1}_{\theta} \leftarrow \alpha^{i}_{\theta} - (1- \gamma)\nabla_{\alpha^{i}_{\theta}} \mathcal{L}_{val}(\alpha^{i}_{\theta}, w*) - \gamma\nabla_{\alpha^{i}_{\theta}}N(\alpha^{i}_{\theta}, A),
\end{equation}
where $N(\alpha^{i}_{\theta}, A)$ is a measure of architecture $\alpha^{i}_{\theta}$ from the history of architectures $A$. This enhancement led to a higher probability of sampling novel architectures rather than well-trained architectures from previous iterations.

\textbf{DARTS-} \cite{chu2020darts} tackled the global performance collapse induced by \textit{skip connections} by adding an auxiliary \textit{skip connection} to the classic mixed output of operations (see Eq. \ref{eq:mixed_output}). The authors asserted that previous works based on analyzing the Hessian matrix eigenvalues (e.g., R-DARTS \cite{Zela2020Understanding}) were imperfect as those methods tend to reject good architectures if they do not meet some arbitrary threshold. This auxiliary operation is pondered by $\beta$, a coefficient independent from architectural weights that is progressively decayed to 0 during the search phase. Moreover, the authors introduced $\beta^{skip}$, a parameter that denotes the importance of \textit{skip connections} inside of the mixed output of operations. Thus Eq. \ref{eq:mixed_output} is modified as follows:
\begin{equation}
    \overline{o}_{i,j}(x) = (\beta + \beta_{i,j}^{skip})x+\sum_{o \in O \setminus \{skip\}} \frac{exp(\alpha^{o}_{i,j})}{\sum_{o'\in O \setminus \{skip\}} exp(\alpha^{o'}_{i,j})}o_{i,j}(x).
\label{eq:dart_aux}
\end{equation}
Fig. \ref{fig:beta_darts} (b) illustrates this mechanism. The authors showed that DARTS- can significantly improve robustness and stabilization during the search process, with +0.5 \% improvement on CIFAR-10 \cite{krizhevsky2009learning}, and +4.5 \% on ImageNet \cite{deng2009imagenet} compared to standard DARTS. DARTS- also uses fewer computational resources than previous approaches such as R-DARTS \cite{Zela2020Understanding}. In addition to standard DARTS, this approach is able to improve the performance of other derivatives such as P-DARTS \cite{chen2019progressive} or PC-DARTS \cite{xu2019pc}.

Path-Regularized Differential Network Architecture Search (\textbf{PR-DARTS}) \cite{zhou2020theory} was the first method to propose a theoretical in-depth analysis of why the over-representation of \textit{skip connections} phenomena happens and how it is connected to performance collapse. This differs from prior works \cite{chu2020fair, chen2019progressive, liang2019darts+} that mainly observed this issue and empirically tested their own solutions. In particular, the authors introduced a convergence theorem demonstrating that the number of \textit{skip connections} heavily influences the supernet's convergence rate (i.e., the more \textit{skip connections}, the faster the supernet converges). This is linked to \textit{skip connections} faster decaying the validation loss $\mathcal{L}_{val}$, and thus leading DARTS search algorithm to increase $\alpha$ weights associated with \textit{skip connections} at the cost of decreasing all other weights. To palliate this issue, they replaced architectural weights with stochastic binary gates, denoted $g_{i,j}^{k}$ for the kth operation between nodes $i$ and $j$. At each iteration, $g_{i,j}^{k}$ is sampled from a Bernoulli distribution to compute the output of each node. Thus, Eq. \ref{eq:mixed_output} is modified as follows:
\begin{equation}
    \overline{o}_{i,j} = \sum_{k=1}^{K} g_{i,j}^{k} o^{k}_{i,j}(x).
\end{equation}
However, leaving the gates unregularized could bias the operation selection in cells since DARTS will increase the weights of all operations to achieve faster convergence. Furthermore, increasing the value of any operation weight could reduce or maintain the loss $\mathcal{L}_{val}$. The authors resolve these issues by using a group-structured sparsity regularization on the gates via rescaling and imposing thresholds: $g_{i,j}^{k} = min(1, max(0, a+(b-a))\overline{g}_{i,j}^{k})$ with $a < 0$ and $b > 1$, and $\overline{g}_{i,j}^{k}$ is an approximation of $g_{i,j}^{k}$ using the Gumbel reparametrization trick. This regularization is expressed by two loss functions targeting \textit{skip} and \textit{non-skip} connections respectively:
\begin{equation}
    \mathcal{L}_{skip}(\alpha) = \zeta \sum_{l=1}^{h-1} \sum_{s=0}^{l-1} \sigma(\alpha_{s,l}^{skip}-\tau\text{log}\left( \frac{-a}{b}\right)), \mathcal{L}_{non-skip}(\alpha) = \frac{\zeta}{r-1} \sum_{l=1}^{h-1} \sum_{s=0}^{l-1} \sum_{k=1}^{K} \sigma(\alpha_{s,l}^{k}-\tau\text{log}\left( \frac{-a}{b}\right)),
\end{equation}
where $\sigma$ denotes the sigmoid function, $h$ is the path depth, $\zeta = \frac{2}{h(h-1)}$ and $\tau$ is a temperature hyperparameter.
In addition, they introduced path regularization to reduce the unfair competition between deep and shallow cells (i.e., a cell containing a large amount of intermediary \textit{skip connections}) as follows:
\begin{equation}
    \mathcal{L}_{path}(\alpha) = \prod_{l=1}^{h-1} \sum_{k \in O_p} \sigma(\alpha_{l, l+1}^{k} - \tau \text{log}\left( \frac{-a}{b}\right)),
\end{equation}
where $O_p$ denotes the parameterized operations. Hence, Eq. \ref{eq:bilevel_optimization} is modified as follows:
\begin{equation}
    \begin{split}
    \underset{\alpha}{\text{min }} \mathcal{L}_{val}(w^*(\alpha), \alpha)+\lambda_1\mathcal{L}_{skip}(\alpha)+\lambda_2\mathcal{L}_{non-skip}(\alpha)-\lambda_3\mathcal{L}_{path}(\alpha),\\
    \text{s.t. } w^{*}(\alpha) = \underset{w}{\text{argmin }} \mathcal{L}_{train}(w,\alpha),
    \end{split}
\end{equation}
where $\lambda_1$, $\lambda_2$ and $\lambda_3$ are constants.
All of these improvements make PR-DARTS search for performance-oriented networks rather than fast-convergence-oriented ones. Empirical results show that PR-DARTS overperformed DARTS and earlier variants on image classification datasets (ImageNet, CIFAR-10) by a large margin.

The authors \textbf{DARTS+PT} \cite{wang2021rethinking} (presented in Section \ref{sec:gradient_approximation}) showed that their proposed perturbation-based architecture selection method prevents \textit{skip connections} from becoming dominant. From a theoretical aspect, they refer to Greff et al. \cite{greff2016highway}, who proved that ResNet \cite{he2016deep} layers are robust to reordering as their outputs correspond to the same estimated optimal feature map values. As the presence of \textit{skip connections} makes DARTS' supernet resembles ResNet, this may explain why DARTS layers are also robust to reordering. Thus, this fact indicates that edges in a cell all try to estimate the same optimal feature maps $m^*$. Wang et al. \cite{wang2021rethinking} used this finding to define the estimated optimal feature maps $\bar{m}_e$ for input $x_e$ of edge $e$:
\begin{equation}
    \begin{split}
    \bar{m}_e(x_e) = \frac{exp(\alpha_{conv})}{exp(\alpha_{conv})+exp(\alpha_{skip})}o_{e}(x_e) + \frac{exp(\alpha_{skip})}{exp(\alpha_{conv})+exp(\alpha_{skip})}x_e\\
    \text{with } \alpha_{conv} \propto (x_{e}-m^{*}), \alpha_{skip} \propto (o_{e}(x_{e})-m^{*})
    \end{split},
    \label{eq:skip_connect}
\end{equation}
where $\alpha_{conv}$ and $\alpha_{skip}$ are architectural parameters, and $o_e$ is the mixed output of operations associated with edge $e$. It can be deduced from Eq. \ref{eq:skip_connect} that the better the supernet is optimized, the closer $x_e$ will get to $m^*$ (since the goal of the training phase is to make edges estimate $m^*$). Consequently, this will widen the $(\alpha_{skip}-\alpha_{conv})$ gap and ultimately this will lead to $\alpha_{skip} > \alpha_{conv}$. However, Wang et al. showed that this only becomes problematic if the architecture selection process relies on $\alpha$. On the contrary, DARTS+PT does not suffer from this issue, although it retains the same search algorithm as DARTS.

Another work dubbed \textbf{NoisyDARTS} \cite{chunoisy2021} addressed this problem in an original way. The authors proposed to inject unbiased random noise during training to prevent the optimizer from increasing architectural weights associated with \textit{skip connections} ($\alpha_{skip}$) too much. They argued that adding noise is an efficient way to improve generalization by smoothing the loss landscape, as pointed out by a prior study \cite{wu2020revisiting}. In practice, NoisyDARTS adds Gaussian noise to the input of \textit{skip connections}. Thus, Eq. \ref{eq:mixed_output} can be rewritten as 
\begin{equation}
    \overline{o}_{i,j}(x) = \sum^{K-1}_{k=1} \text{softmax}(\alpha^{k}_{i,j})o^{k}_{i,j}(x) + \text{softmax}(\alpha^{\text{skip}}_{i,j})o^{\text{skip}}_{i,j}(x+\tilde{x}),
\end{equation}
where $\tilde{x} \sim \mathcal{N}(\mu, \sigma^2)$ is a random noise sampled from a Gaussian distribution parameterized by mean $\mu$ and variance $\sigma^2$ ($\mu$ = 0 and $\sigma=0.2$ when searching on ImageNet). Despite its simplicity, NoisyDARTS managed to suppress \textit{skip connections} and consistently overperformed prior DARTS derivatives on CIFAR-10 \cite{krizhevsky2009learning}, ImageNet \cite{deng2009imagenet}, and NAS-Bench-201 \cite{dong2020nasbench201} (e.g., +9.7 \% top 1 accuracy on ImageNet compared to DARTS).

Ye et al. (\textbf{$\beta$-DARTS}) \cite{ye2022b} continued the work initiated by DARTS- \cite{chu2020darts} with the introduction of the Beta-Decay regularization method (presented in detail in Section \ref{sec:gradient_approximation}). In addition to reducing the optimization gap problem, they argued that the Beta-Decay mechanism also alleviates the over-representation of \textit{skip connections} issue and ensures fair competition between the operations. More accurately, the authors provide a theoretical explanation with the following equation:
\begin{equation}
    \phi \propto \sum_{i=0}^{h-2}[(\theta_{i,h-1}^{conv}\beta_{i,h-1}^{conv})^{2}\prod_{t=0}^{i-1}(\theta_{i,h-1}^{skip}\beta_{i,h-1}^{skip})],
    \label{eq:beta_decay}
\end{equation}
where $h$ is the number of layers in the supernet, $\phi$ represents the architectural weight gradients, $\theta$ represents the influence of the Beta Decay regularization, and $\beta_{k} = \frac{exp(\alpha_k)}{\sum_{k'\in O}exp(\alpha_{k'}} = softmax(\alpha_k)$.
By taking into account that $\theta$ varies antagonistically to $\beta$ (as it is a regularization function),  Eq. \ref{eq:beta_decay} shows us that the convergence of networks rely more on $\beta_{conv}$ than on $\beta_{skip}$. Consequently, this means that Beta-Decay helps to reduce the prominence of \textit{skip connections}.

\textbf{CDARTS} \cite{yu2022cyclic} straightforwardly addressed the over-representation issue. They simply added a L1 regularization factor to the architectural weights of non-parametric operations $O_{np} = {skip\_connect, max\_pool\_3x3, avg\_pool\_3x3}$ as:
\begin{equation}
    \mathcal{L}_{reg} = \lambda\sum_{o\in O_{np}}|\alpha_{o}|,
\end{equation}
where $\lambda$ is a hyperparameter that balances the value of $\mathcal{L}_{reg}$. The authors showed that this method successfully prevented the operations in $O_{np}$ from becoming dominant.

\subsection{Computational Efficiency and Latency Reduction}
\label{sec:computational_efficiency}

Chen et al. \cite{chen2019progressive} defined the problem of \textit{NAS in the wild} as being able to search for an architecture on a proxy dataset (e.g., CIFAR-10 \cite{krizhevsky2009learning}) to limit computational cost and successfully transfer to another, more challenging dataset (e.g., ImageNet \cite{deng2009imagenet}). Most DARTS derivatives followed this paradigm, contrary to most non-DARTS approaches such as ProxylessNAS \cite{cai2018proxylessnas}.

As discussed above (see Section \ref{sec:gradient_approximation} for additional details), \textbf{ProxylessNAS} searches directly on the target dataset (e.g., ImageNet \cite{deng2009imagenet}) rather than a proxy dataset (e.g., CIFAR-10). It also enforces latency constraints on specific hardware (e.g., mobile phones, GPU, or CPU). Hence, it is a multi-objective NAS method, but one of the objectives (latency) is not differentiable. Instead, latency is measured in real-time on GPU/CPU and is predicted from a lookup table on mobile settings. ProxylessNAS successfully constrained mobile latency to a similar level to MobileNetV2 \cite{sandler2018mobilenetv2} and improved top 1 accuracy by 2.6 \% on ImageNet. 

\begin{figure}[h]
  \centering
  \includegraphics[width=0.65\linewidth]{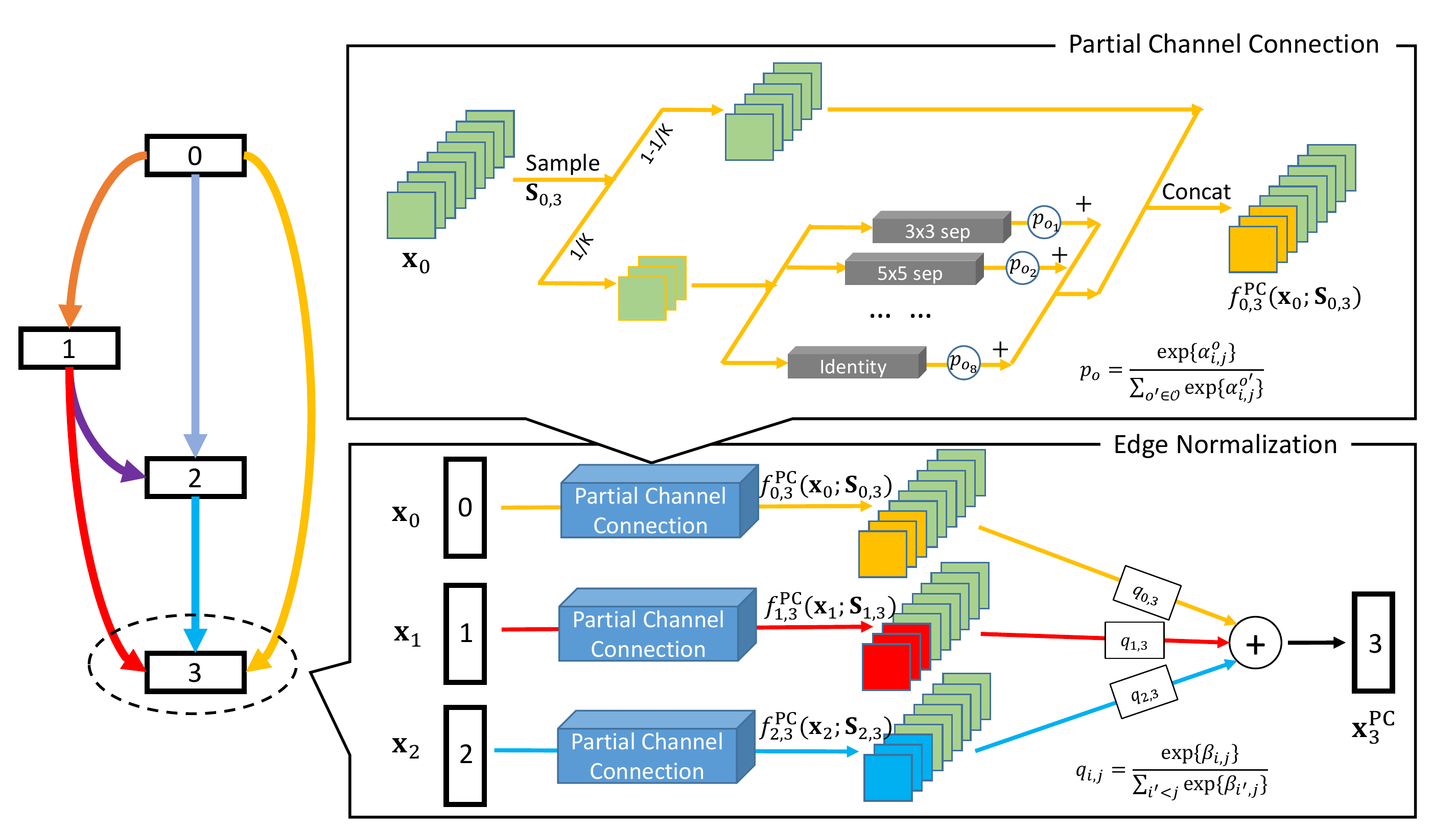}
  \caption{Layout of the PC-DARTS search process. Figure from Xu et al. \cite{xu2019pc}}
  \label{fig:pcdarts}
\end{figure}

With \textbf{PC-DARTS} (Partially-Connected DARTS), Xu et al. \cite{xu2019pc} sought to improve computational efficiency without compromising performance. To this end, they perform architecture search in only a subset of randomly sampled channels while bypassing the rest. This concept is based on the assumption that computation on this subset is an adequate approximation of the effective computation on all the channels. Considering edge $e_{i,j}$, partial channel connection involves defining a channel sampling mask $S_{i,j}$ which nullifies (i.e., assigns a weight value of 0) all channels except selected ones in the mixed output $\overline{o}_{i,j}$, thus modifying Eq. \ref{eq:mixed_output} as follows:
\begin{equation}
    \overline{o}^{PC}_{i,j}(x) = \sum^{K}_{k=1} \frac{\text{exp}(\alpha^{k}_{i,j})}{\sum^{K}_{k'=1} \text{exp}(\alpha^{k'}_{i,j})}o^{k}_{i,j}(S_{i,j}x) + (1-S_{i,j})x.
\label{eq:pc_darts}
\end{equation}
This process has the advantage of reducing the memory overhead by $K$ times, with $\frac{1}{K}$ being the channel selection ratio. Subsequently, it helps reduce the search cost on CIFAR-10 from 1 GPU day (DARTS) to only 0.1 GPU day, and PC-DARTS achieves a better top 1 accuracy on ImageNet than ProxylessNAS (75.8 \% vs.  75.1 \%) with half the search cost. However, partial channel connection induces an inconsistency in the selection of operations across the different sampled channels. To palliate this issue, the authors introduced an additional set of learning parameters $\beta_{i,j}$ that are shared throughout the search process to act as an \textit{edge normalization} mechanism. The PC-DARTS approach is summarized in Fig \ref{fig:pcdarts}.

Cai et al. proposed Once-for-All (\textbf{OFA}) \cite{cai2020once} as a solution for decoupling the training and search phases with the objective of drastically reducing the computational cost of DNAS. In particular, they trained a single large supernet whose configuration (e.g., kernel size, depth, or width) can be altered and directly deployed without further training. OFA follows two stages: (1) a training phase where the different subnetworks that compose the supernet are optimized to improve their accuracy (2) a hardware-aware NAS phase (model specialization stage) where sub-networks are sampled to train accuracy and latency predictors. This enables OFA to target specific hardware and latencies. However, since simultaneously optimizing the parameters of the very large ($10^{19}$) number of subnetworks is prohibitively expensive, the authors introduced a novel training process during which the OFA network is progressively fine-tuned to train subnetworks of increasingly smaller size. This is akin to a pruning process performed over different modalities (i.e., input resolution, width, depth, and kernel size). Overall, OFA only requires $4.2k$ GPU hours for end-to-end training (3 times lower than DARTS \cite{liu2018darts}) to overperform all previous approaches on ImageNet in the mobile setting (i.e., less than 600M FLOPS). 

Although it was not their primary objective, the authors of \textbf{DOTS} \cite{gu2021dots} were able to reduce computational cost to only 0.26 GPU day when searching on CIFAR-10 and 1.3 GPU days when searching on ImageNet. This is due to the decoupling between the topology search and operation search that greatly reduces the number of candidate operations on each edge (and thus the search space size), making both processes converge fast.

Wu et al. \cite{wu2019fbnet} proposed \textbf{FBNet} as a DNAS framework aimed at improving latency and computational efficiency, especially targeted at low-power hardware such as mobile phones. Firstly, FBNet browses a search space $\mathcal{A}$ different than DARTS' that is not organized around cell building blocks but rather around layers. The macro-architecture (i.e., the pre-processing/post-processing layers, the number of intermediate layers, and their input shapes) is fixed, whereas independent architectures (from a selection of "blocks") are searched for each layer. This leads to a greater diversity of candidate architectures than in DARTS' search space. In addition, FBNet combines the standard Cross-Entropy loss $\mathcal{L}_{CE}$ with a hardware-aware latency loss $\mathcal{L}_{LAT}$ defined as follows:
\begin{equation}
    \mathcal{L}_{LAT} = \alpha \text{log}\left(\sum_{i} LAT(b_{l}^{a})\right)^\beta,
\end{equation}
where $LAT(b_{l}^{a})$ denotes the latency of block $a$ of the $l$-th layer, and $\alpha$ and $\beta$ are coefficients weighting $\mathcal{L}_{LAT}$. The latency values are retrieved from a latency lookup table (similarly to ProxylessNAS \cite{cai2018proxylessnas}), as measuring latency from mobile processors in real time is prohibitively expensive. In addition, using a lookup table makes $\mathcal{L}_{LAT}$ differentiable. Finally, the authors devised a differentiable NAS algorithm where the search space is modeled by a stochastic supernet. Thus, only one candidate block is sampled at a time independently for each layer from a probability distribution obtained through a softmax instead of a weighted mixed output of operation as featured in DARTS (see Eq. \ref{eq:mixed_output}). Consequently, the output $x_l$ of layer $l$ is a masked output defined as:
\begin{equation}
    x_l = \sum_i m_{l,i}b_{l,i}(x_{l-1}),
\end{equation}
where $m_{l,i}$ is a mask that equals to 1 if block $b_{l,i}$ is sampled or 0 otherwise. Therefore, the probability of sampling an architecture $a \in \mathcal{A}$ is described by the following equation:
\begin{equation}
    P_{\Theta}(a) = \prod_{l} P_{\Theta_l}(b_{l}=b_{l,i}^{a}),
\end{equation}
where $\Theta$ is composed of all the parameters that determine the sampling probabilities of blocks for each layer. Furthermore, the authors resorted to the Gumbel-Softmax \cite{jang2017categorical} function to relax the discrete masks $m_l$ into a continuous distribution and thus make the whole search process differentiable w.r.t. the sampling parameters $\Theta$. The authors empirically showed that FBNet reached a higher top 1 accuracy (e.g., +1.8 \% for FBNet-C) on ImageNet than DARTS for a 33 \% lower search cost. FBNet-A also reached a latency as low as 19.8 ms when targeting a Samsung Galaxy S8.

However, FBNet is not free from limitations, and hence Wan et al. \cite{wan2020fbnetv2} designed an updated method dubbed \textbf{FBNetV2}. Their primary concern was to palliate the small search space size issue present in FBNet and DARTS. Consequently, they introduced a greatly enlarged search space (see Section \ref{sec:search_space}). To keep their method computationally efficient, the authors proposed DMaskingNAS. This NAS algorithm uses weight-sharing approximations to efficiently search over additional hyperparameters, such as the number of filters and the input dimensions. They kept the layer-wise DNAS paradigm described in FBNet but used a channel-masking mechanism parameterized through a Gumbel-Softmax function. Thus the output $y$ of a block $b$ can be computed as follows:
\begin{equation}
    y = b(x) \circ \sum_{i=1}^{k} g_i \mathbbm{1}_i,
\end{equation}
where $g_i$ denotes Gumbel weights and $\mathbbm{1}_i$ is a mask vector whose first $i$ values are 1s with the rest being 0s. This way, each block's channel number can be searched without significant additional computational cost. Furthermore, FBNetV2 searches for different input resolutions by performing resolution subsampling from the original input (i.e., extracting smaller input feature maps using the nearest neighbors method). Once the output feature map has been computed, it is upsampled into a larger fixed-size one to preserve dimensional consistency. FBNetV2 maintains a computational cost similar to FBNet despite searching on a search space up to $10^{14}$ times larger.

\textbf{VIM-NAS} (Variational Information Maximization Neural Architecture Search) \cite{wang2021learning} observed that each cell edge is considered independent in the global architecture of previous DNAS methods. In contrast, the authors introduced a novel way of formulating the NAS problem by assuming that the architectural distribution $\mathcal{A}$ is a latent representation of specific data points from dataset $\mathcal{D}$ such as there is a distribution $p_{\phi}(\mathcal{D}, \mathcal{A}) = p(\mathcal{D})p_{\phi}(\mathcal{A}|\mathcal{D})$ parameterized by $\phi$. More specifically, VIM-NAS strives to maximize the mutual information $I_{\phi}(\mathcal{D}, \mathcal{A})$ between $\mathcal{A}$ and $\mathcal{D}$ as
\begin{equation}
    \underset{\phi}{\text{max }} I_{\phi}(\mathcal{D}, \mathcal{A}) = \mathbb{E}_{p_{\phi}(\mathcal{D}, \mathcal{A}} [\text{log}p_{\phi}(\mathcal{D}|\mathcal{A})].
\end{equation}
Thus, the objective $\mathcal{L}(\phi,\theta,\mathcal{D})$ of the DNAS process can be formulated as
\begin{equation}
    \underset{\theta, \phi}{\text{max }} \mathcal{L}(\phi,\theta,\mathcal{D}) = \sum_{d \in \mathcal{D}} \mathbb{E}_{p_{\phi}(\mathcal{D}, \mathcal{A}} [\text{log}q_{\theta}(\mathcal{D}, \mathcal{A}],
\end{equation}
where $\text{log}q_{\theta}(\mathcal{D}, \mathcal{A})$ is a supernet approximation of $\text{log}p_{\phi}(\mathcal{D}|\mathcal{A})$. In practice, $p_{\phi}(\mathcal{D}|\mathcal{A})$ is reformulated to the Gaussian noise $\mathcal{N}(\mu_{\theta},1)$ where $\mu_{\phi}$ is parameterized by the convolutional network $\phi$. This makes VIM-NAS very fast as it can converge in only a single epoch in DARTS' search space (i.e., a 0.007 GPU day search cost) while providing a top-1 accuracy improvement of +0.55 \% on CIFAR-10 \cite{krizhevsky2009learning} and +2.04 \% on ImageNet \cite{deng2009imagenet} compared to DARTS.

\begin{figure}[h]
  \centering
  \includegraphics[width=0.55\linewidth]{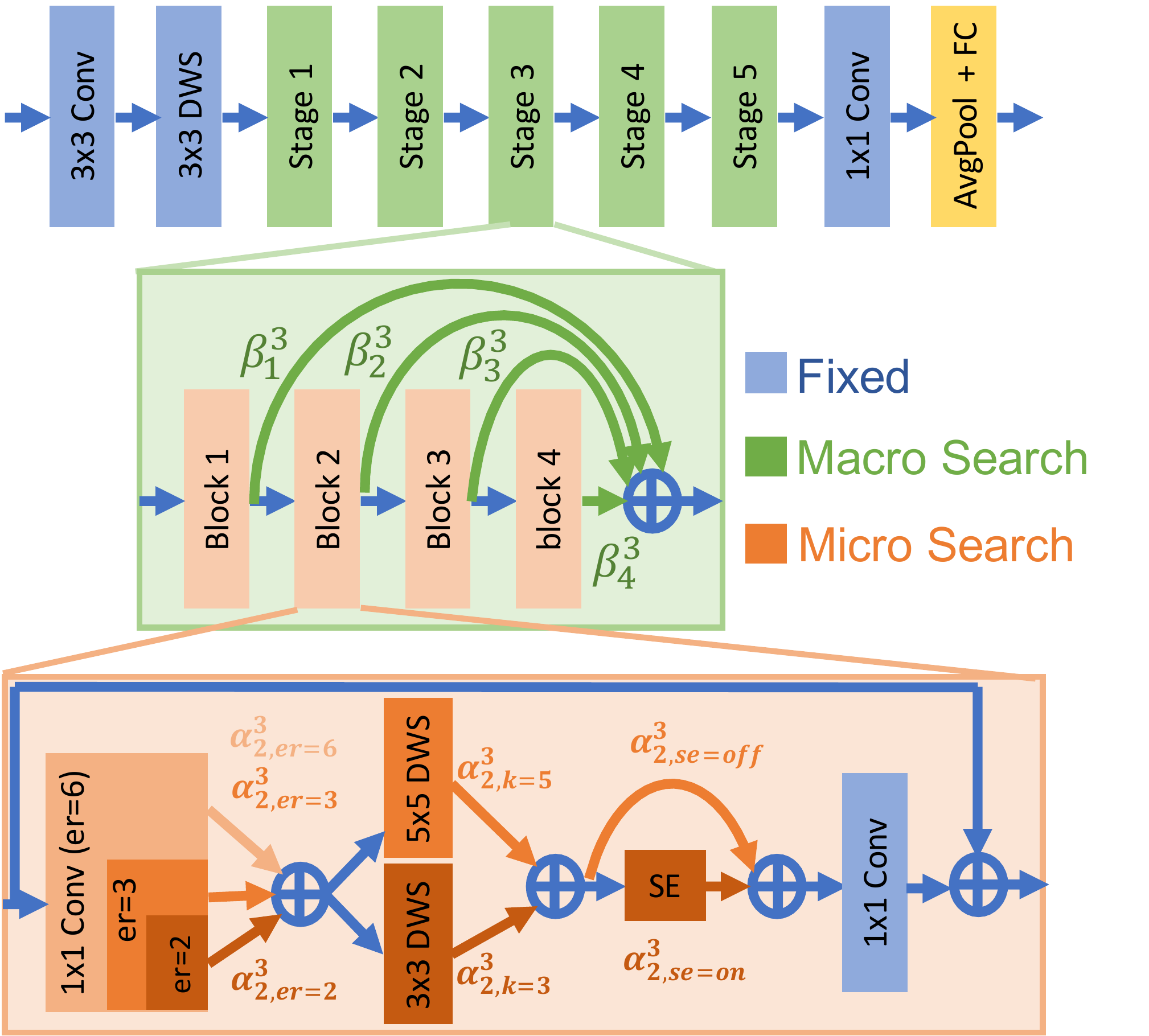}
  \caption{Layout of the HardCoRe-NAS search process. Figure from Nayman et al. \cite{nayman2021hardcore}}
  \label{fig:hardcorenas}
\end{figure}

Methods such as FBNet \cite{wu2019fbnet} or ProxylessNAS \cite{cai2018proxylessnas} sought to impose hardware and latency constraints softly by formulating an objective function which is a trade-off between accuracy and computational resources. In contrast, \textbf{HardCoRe-NAS} (Hard Constrained diffeRentiable NAS) \cite{nayman2021hardcore} searches for high-accuracy architectures that strictly respect a hard latency constraint. The authors reformulated the classic bilevel optimization problem of DNAS (see Eq. \ref{eq:bilevel_optimization}) as
\begin{equation}
\begin{split}
    \underset{\zeta \in S}{\text{min }}\mathbb{E}_{x,y~\mathcal{D}_{val};\hat{\zeta}~\mathcal{P}_{\zeta}(\mathcal{S})}[\mathcal{L}_{CE}(x,y|w^{*},\hat{\zeta})] \text{ s.t. } \text{LAT}(\zeta) \leq T\\
    w^{*} = \underset{w}{\text{argmin }}\mathbb{E}_{x,y~\mathcal{D}_{train};\hat{\zeta}~\mathcal{P}_{\zeta}(\mathcal{S})}[\mathcal{L}_{CE}(x,y|w,\hat{\zeta})],
\end{split}
\end{equation}
where $\mathcal{S}$ is a fully differentiable block-based search space parameterized by $\zeta = (\alpha, \beta)$, $\mathcal{D}_{train}$ and $\mathcal{D}_{val}$ are the train and validation datasets' distributions, $\mathcal{P}_{\zeta}(\mathcal{S})$ is a probability measure over $S$, and $\text{LAT}(\alpha, \beta)$ is the estimated latency of the model. $\mathcal{S}$ is composed of a micro $\mathcal{A}$ (i.e., block internal architecture $c \in \mathcal{C}$) and macro (i.e., connections between blocks at every stage $s \in S$) $\mathcal{B}$ search spaces parameterized by $\alpha \in \mathcal{A}$ and $\beta \in \mathcal{B}$ respectively. Thus, the overall expected latency $\text{LAT}(\alpha, \beta)$ is computed by summing over the latency $t^{s}_{b,c}$ for every possible configuration $c\in\mathcal{C}$ of every block $b$, over all possible depths $d$, and over all the stages:
\begin{equation}
    \text{LAT}(\alpha,\beta) = \sum_{s=1}^{S} \sum_{b'=1}^{d} \sum_{b=1}^{b'} \sum_{c\in\mathcal{C}} \alpha^{s}_{b,c} \cdot t^{s}_{b,c} \cdot \beta^{s}_{b'}. 
\end{equation}
HardCoRe-NAS uses $\text{LAT}$ to build a constrained search space $\mathcal{S}_{LAT}= \{\zeta|\zeta\in\mathcal{P}_{\zeta}(\mathcal{S}), \text{LAT}(\zeta)\leq T\}$. Similarly to DARTS, $\mathcal{S}$ is relaxed to be continuous by searching for $\zeta \in \mathcal{S}_{LAT}$. The search process of HardCoRe-NAS is visually summarized in Fig \ref{fig:hardcorenas}.
The authors experimentally showed that HardCoRe-NAS managed to constrain latency to the same level or lower than previous methods while overperforming them on ImageNet \cite{deng2009imagenet} (e.g., -9 ms latency reduction and +1.6 \% top 1 accuracy improvement compared to FBNet on an Nvidia P100 GPU).

Similarly to UNAS \cite{Vahdat_2020_CVPR}, \textbf{RADARS} \cite{yan2022radars} (Reinforcement Learning Aided Differentiable Architecture Search) leverages Reinforcement Learning to help the differentiable search process. However, RADARS focuses on reducing computational and memory costs, whilst UNAS is performance-oriented. The RL algorithm prunes the search space through iterative exploration/exploitation phases. It identifies promising subsets of operations for each laver and prunes the search space of the other operations (exploration phase). Differentiable NAS is then performed on this reduced search space instead of the entire search space (exploitation phase). The authors bounded the GPU memory usage to a maximum of 12 Go (Nvidia RTX 2080ti). They showed that RADARS could reach competitive scores for restricted memory (11 Go) and time (3.08 GPU days) on ImageNet despite using a large MobileNet-like search space.  



\subsection{Search Space Restrictions}
\label{sec:search_space}
Bypassing the search space restrictions of DNAS (especially concerning DARTS) is one of the main goals pursued by researchers in the field. For instance, \textbf{ProxylessNAS} \cite{cai2018proxylessnas} step out of the cell-based paradigm to instead search directly for an entire architecture. However, as discussed in Section \ref{sec:gradient_approximation}, this led to an exponential increase in computational resources that the authors alleviated by using a binarization mechanism to only instantiate a single path in memory at a given time.

\textbf{D-DARTS} \cite{heuillet2021d} kept the cell-based paradigm of DARTS but extended it to cover an entire architecture \textit{via} a distributed process. The authors still used two types of cells (\textit{normal} and \textit{reduction}), but they were individualized to each cover a specific portion of the architecture. Each individual cell can be considered a small independent neural network. This means that a search model of size $N$ is now composed of $N$ individual cells $C$ (with \textit{reduction} cells positioned at the 1/3 and 2/3 of the network). To help with the optimization of individualized cells, D-DARTS introduced a novel loss function dubbed \textit{ablation loss} and denoted $\mathcal{L}_{AB}$. This loss function, derived from Game Theory (i.e., from the concept of Shapley Values \cite{shapley1997value}), considers the Neural Architecture Search process as a cooperative game with cells being assimilated to players. Hence, $\mathcal{L}_{AB}$ computes the marginal contribution $M_{C}$ of each cell to the common reward (the validation loss $\mathcal{L}_{val}$) and quantifies the difference between the marginal contribution $M_{C}^{i}$ of given cell $C^{i} \in C$ and the mean marginal contribution:
\begin{equation}
    \mathcal{L}_{AB}^{i} = 
        \begin{cases}
            \frac{M_{C}^{i} - mean(M_C)}{mean(M_C)} & \quad \text{if } mean(M_C) \neq 0 \\
            0 & \quad \text{else}
        \end{cases}
\end{equation}
with $M_{C}^{i}$ defined as follows:
\begin{equation}
    M_{C}^{i} = \mathcal{L}_{val}^{(C)} - \mathcal{L}_{val}^{(C \setminus \{C_{i}\})},
\end{equation}
Furthermore, this distributed search space makes it possible to directly use existing handcrafted architectures as starting points for the search process. A handcrafted architecture can be considered a local minimum in the search space. This way, architectures such as ResNet50 \cite{he2016deep} or Xception \cite{chollet2017xception} have been successfully encoded in D-DARTS' space and have been significantly improved.

\begin{figure}[h]
  \centering
  \includegraphics[width=0.75\linewidth]{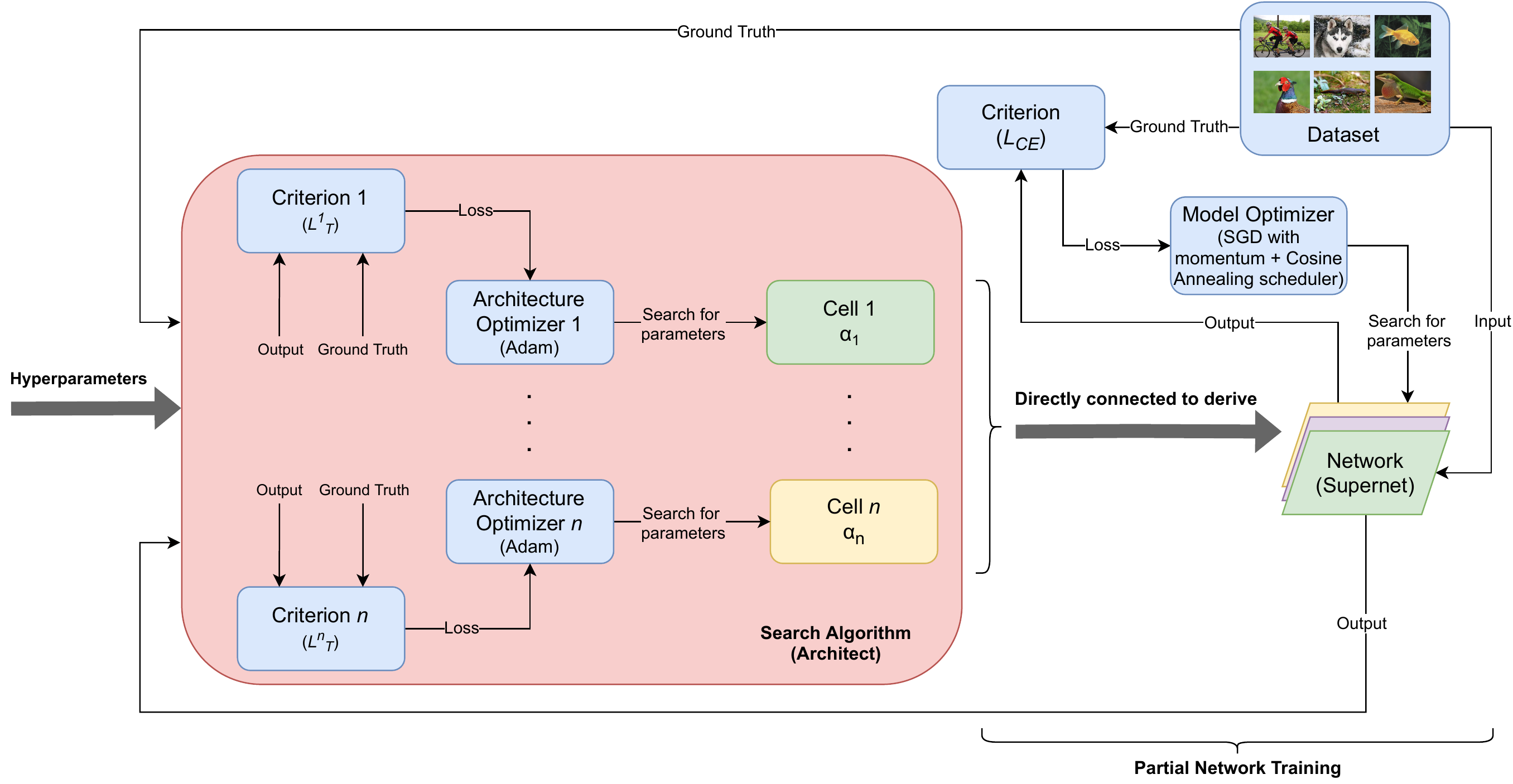}
  \caption{Layout of the D-DARTS search process. Figure from Heuillet et al. \cite{heuillet2021d}.}
  \label{fig:ddarts}
\end{figure}

This process led to a drastic increase in the diversity of discovered architectures that allowed D-DARTS to reach competitive scores in both small-scale (CIFAR-10 \cite{krizhevsky2009learning}) and large-scale (ImageNet \cite{deng2009imagenet}) image classification datasets. The search process of D-DARTS is summarized in Fig. \ref{fig:ddarts}.

The authors of DenseNAS \cite{fang2020densely} introduced a densely-connected block-based search space that allows them to search for block widths and the number of blocks per layer. They designed routing blocks with gradually increasing widths, and each block output linked to multiple other blocks, covering a large spectrum of block widths and block numbers per layer. Each routing block comprises several shape-alignment layers and several basic layers, which are mixed outputs of the different candidate operations. The basic layers are relaxed into continuous operations using \textit{softmax}, similarly to DARTS (see Eq. \ref{eq:mixed_output}). In the same way, the routing block output is a mixed output of all the possible paths leading to the next routing block and relaxed using a \textit{softmax}. Furthermore, DenseNAS features a chained cost estimation algorithm that aims to restrict the computational cost and latency of the final architecture. The cost of each basic layer operation is retrieved from a lookup table and is used to estimate the cost of the whole architecture as follows:
\begin{align}
    cost^l &= \sum_{o\in O} w^{l}_{o} cost^{l}_{o}\\
    \Tilde{cost}^i &= cost^{i}_{b} + \sum_{j=i+1}^{i+m}p_{i,j}(cost^{i,j}_{align} + cost^{j}_{b}),
\end{align}
where $w^{l}_{o}$ is the \textit{softmax} weight of operation $o$ in basic layer $l$, $p_{i,j}$ is the \textit{softmax} weight of the path from routine block $B_i$ to routine block $B_j$, m is the number of subsequent blocks, and $cost^{i,j}_{align}$ is the cost of the shape-alignment layer in block $B_j$ with input from block $B_i$. This cost is then integrated into the loss function
\begin{equation}
    \mathcal{L}(w,\alpha, \beta) = \mathcal{L}_{CE} + \lambda\text{log}_{\mathcal{T}}(\Tilde{cost}^1).
\end{equation}
DenseNAS overperformed previous methods on ImageNet (+2.8 \% top 1 accuracy compared to DARTS) while being able to constrain the number of FLOPS and the latency.

The authors of \textbf{FBNetV2} \cite{wan2020fbnetv2} (first presented in Section \ref{sec:computational_efficiency}) pointed out that the main issue in previous DNAS works (DARTS \cite{liu2018darts} and FBNet \cite{wu2019fbnet} primarily) is related to their severely restricted search space. Therefore, they crafted a novel search space encompassing two new hyperparameters (i.e., the number of channels and the input resolution). This novel search space comprises $10^{35}$ candidate architectures and is $10^{14}$ times larger than FBNet's. By leveraging this expanded search space, FBNetV2 overperforms FBNet on ImageNet significantly (e.g., +1.9 \% top 1 accuracy for FBNetV2-F4 compared to FBNet-B).

\begin{figure}[h]
  \centering
  \includegraphics[width=\linewidth]{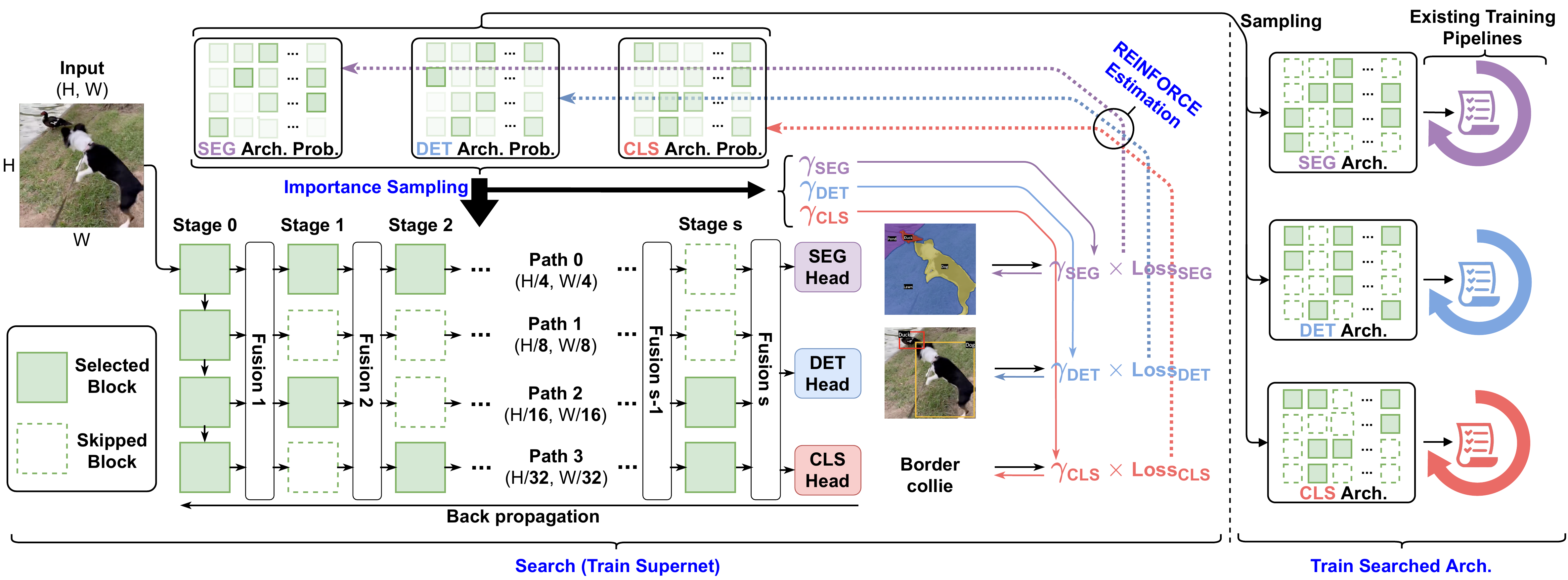}
  \caption{Layout of the FBNetV5 search process. Figure from Wu et al \cite{wu2021fbnetv5}.}
  \label{fig:fbnetv5}
\end{figure}

Building upon what has been laid by previous FBNet works \cite{wu2019fbnet, wan2020fbnetv2, dai2021fbnetv3}, \textbf{FBNetV5} \cite{wu2021fbnetv5} is an interesting work as it did not simply seek to lift search space restrictions but more specifically focused on a less-explored subject: improving the transferability of DNAS architectures between different computer vision tasks. The authors addressed this challenge by combining a differentiable NAS process with FBNetV3, a NAS approach that stepped out of DNAS and instead proposed its own novel paradigm where architectures and training hyperparameters ("recipe") are matched to reach optimal performance. More precisely, they created a supernet trained on a multitask dataset (generated from ImageNet \cite{deng2009imagenet}) to disentangle the search process from the training pipeline of the target. FBNetV5 performs topology search to find the optimal backbone architecture for each task (i.e., semantic segmentation, object detection, and image classification). This is done simultaneously for all tasks.
Furthermore, the authors designed a search algorithm that produces architectures for each task at a constant computational agnostic to the number of tasks. At the task level, they leverage a DNAS process following \cite{wu2019fbnet} where they browsed a block-based search space $\mathcal{A} = \{0,1\}^B$ comprising $B$ blocks. An architecture $a \in \mathcal{A}$ is hence a set of binary masks $a_b$ sampled independently from a Bernoulli distribution. When searching on multiple tasks, the DNAS problem is relaxed as follows:
\begin{equation}
    \text{min}_{\pi^1,...,\pi^T,w} \sum_{t=1}^{T} \mathbb{E}_{a^{t}~p_{\pi^t}} \{\mathcal{l}^{t}(a^{t},w)\},
\end{equation}
where $a^t$ are architectures sampled from task-specific distributions $p_{\pi^t}$, $l^t$ are task-specific losses, $T$ is the number of tasks, and $w$ denotes the supernet weights. In addition, to restrict computational cost, the authors adopt the RL algorithm REINFORCE \cite{williams1992simple} and importance sampling \cite{hastings1970monte} to reduce the number of forward and backward passes of the search algorithm from $T$ to 1. Fig. \ref{fig:fbnetv5} summarizes the search process of FBNetV5. FBNetV5 successfully overperforms all previous NAS methods on datasets (ImageNet \cite{deng2009imagenet}, COCO \cite{lin2014microsoft}, and ADE20k \cite{zhou2017scene}) corresponding to the three tasks this method focused on.

\section{Applications}
\label{sec:applications}

One salient fact to note is that most of the major DNAS works (e.g., DARTS \cite{liu2018darts}, ProxylessNAS \cite{cai2018proxylessnas}, or FBNet \cite{wu2019fbnet}) focused on improving CNN architectures targeted at computer vision tasks. Hence, they mostly validate their approaches on image classification datasets (e.g., CIFAR-10/100 \cite{krizhevsky2009learning}, or ImageNet \cite{deng2009imagenet}) and object detection or semantic segmentation datasets (e.g., MS-COCO \cite{lin2014microsoft}, Pascal-VOC \cite{everingham2012pascal}, or Cityscapes \cite{Cordts2016Cityscapes}). However, many works explored other fields of application. In addition to the major approaches discussed in the following paragraphs and Section \ref{sec:review}, we included more minor works that focused on a specific application of DNAS. Table \ref{tab:applications} summarizes those approaches.

\textbf{Remote Sensing} A few applications were developed to address image-oriented tasks such as radar image analysis \cite{dong2020automatic, zhang2021self} and scene classification \cite{peng2020efficient}. For instance, Zhang et al. \cite{zhang2021self} leveraged a DARTS-based method to search for AutoDL detector CNN backbones optimized for sonar image and maritime radar image analysis.  

\textbf{Natural Language Processing.} DARTS \cite{liu2018darts}, and some of its variants (such as $\beta$-DARTS \cite{ye2022b}) searched for Recurrent Neural Networks (RNN) \cite{rumelhart1986learning} architectures to perform Natural Language Processing (NLP) tasks such as language modeling on the Penn Tree Bank \cite{mitchell1993penn} and WikiText-2 \cite{merity2017pointer} datasets. In the same area, AdaBERT \cite{chen2021adabert} leveraged Differentiable NAS to compress BERT \cite{kenton2019bert} large language models into smaller task-oriented models. 

\textbf{Medical Applications.} Numerous works proposed DNAS-based approaches to analyze medical data, such as MRIs \cite{li2021differentiable, lu2022rt} or volumetric images \cite{zhu2019v, yu2020c2fnas, huang2021adwu}. The application to medical imaging is straightforward as most developed DNAS approaches target CNNs and computer vision tasks, as shown in Section \ref{sec:review}. Forr example, Guo et al. \cite{guo2020organ} developed a DNAS approach to modulate the composition of CNNs used to perform organ at risk segmentation in patients treated for head and neck cancer.

\textbf{Reinforcement Learning.} RL-DARTS \cite{miao2021rl} used DARTS to search for backbone architectures for Deep Reinforcement Learning on-policy and off-policy algorithms. They showed that performing Differentiable NAS with DARTS is relevant for both discrete action and continuous control environments. For instance, in the Procgen \cite{cobbe2020leveraging} environment, RL-DARTS improved performance by 250 \% over the baseline IMPALA-CNN \cite{espeholt2018impala}.

\textbf{Audio Processing.} A few studies targeted speech recognition with the aim of discovering novel convolutional neural network architectures specialized in audio pattern extraction \cite{zheng2021efficient, hu2022neural, ding2020autospeech}. For instance, Hu et al. \cite{hu2022neural} designed a DARTS-based search strategy to discover Time Delay Neural Network (TDNN) architectures optimized for automatic speech recognition.

\textbf{Hardware Optimization.} Matching architectures and hardware has been a major focus of DNAS. Multiple works (some, such as ProxylessNAS \cite{cai2018proxylessnas}, are discussed in detail in Section \ref{sec:computational_efficiency}) tried to optimize computational cost and latency on multiple platforms such as GPUs \cite{cai2018proxylessnas, wu2019fbnet, yan2022radars}, CPUs \cite{cai2018proxylessnas}, and embedded systems \cite{cai2018proxylessnas, kim2021mdarts, lopez2021dnas}. In our opinion, this is paramount to ensure the real-world deployment of Deep Learning, as consumer-grade devices are not as powerful as high-end/data-center GPUs/TPUs typically used by researchers.

\begin{table}
  \caption{Summary of Differentiable NAS works according to their field of application. Some articles are referenced under multiple categories.}
  \label{tab:applications}
  \small
\resizebox{\textwidth}{!}{
  \begin{tabular}{lll}
    \toprule
    Field & Subfield & References\\
    \midrule
    \multirow{9}{4em}{Computer Vision} & Image Classification & \cite{liu2018darts, wan2020fbnetv2, wu2019fbnet, wu2021fbnetv5, chu2020fair, zhang2021idarts, liang2019darts+, heuillet2021d, chu2020darts, xu2019pc, chen2019progressive, chen2020stabilizing} \\
    & Object Detection & \cite{marvasti2021chase, li2020differentiable} \\
    & Video Processing & \cite{peng2019video} \\
    & Semantic Segmentation & \cite{zhu2019v, liu2019auto, fan2022self} \\
    & Image Super-Resolution & \cite{huang2021lightweight, weng2021improved, wu2021trilevel} \\
    & Image Denoising & \cite{gou2020clearer, zhang2020memory} \\
    & Pose Estimation & \cite{zhang2021efficientpose} \\
    & Image Generation & \cite{gao2020adversarialnas} \\
    & Facial Recognition & \cite{li2021auto} \\
    \hline
    \multirow{2}{4em}{Remote Sensing 
    } & Radar Image Analysis & \cite{dong2020automatic, zhang2021self}\\
    & Scene Classification & \cite{peng2020efficient} \\
    \hline
    \multirow{3}{4em}{Natural Language Processing} & Language Modeling & \cite{liu2018darts, ye2022b, jiang2019improved}\\
    & Keyword Spotting & \cite{mo2020neural} \\
    & & \\
    \hline
    Medical Applications & Medical Image Analysis & \cite{li2021differentiable, zhu2019v, guo2020organ, lu2022rt, huang2021adwu, weng2019unet, yu2020c2fnas}\\
    \hline
    Reinforcement Learning & Deep Reinforcement Learning Backbone & \cite{miao2021rl}\\
    \hline
    Audio Processing & Speech Recognition & \cite{zheng2021efficient, hu2022neural, ding2020autospeech} \\
    \hline
    \multirow{2}{4em}{Hardware Optimization} & Embedded Systems and Latency Reduction & \cite{lopez2021dnas, cai2018proxylessnas, jiang2021eh, yu2020bignas, wu2019fbnet, yan2022radars, kim2021mdarts, luo2022lightnas, nayman2021hardcore, yuzuguler2022u, cai2020once, peng2022recnas, zhang2020fast}\\
    & Predictive Maintenance & \cite{zhang2021differentiable}\\
  \bottomrule
\end{tabular}}
\end{table}

\section{Discussion and Future Directions}
\label{sec:discussion}
In Section \ref{sec:review}, we reviewed 26 recent DNAS approaches that targeted 4 different challenges (presented in Section \ref{sec:darts}). More than half (62 \%) of these approaches are based on DARTS \cite{liu2018darts} due to the high popularity this method enjoyed from the moment it was first published (2019) until now, with novel DARTS-based methods still being proposed in 2022 \cite{ye2022b, yu2022cyclic}. Each reviewed DNAS method addressed at least one of the four challenges we identified in Section \ref{sec:darts}.

One noteworthy fact (clearly shown in Fig.\ref{fig:taxonomic_tree}) is that there is a clear partition between DARTS-based and non-DARTS-based works when considering the challenges they tackled. The vast majority (81 \%) of DARTS-based methods addressed either challenge \textbf{I} (gradient approximation discrepancies) or challenge \textbf{II} (over-representations of \textit{skip connections}) while the rest mostly targeted challenge \textbf{III} (computational efficiency and latency reduction) and challenge \textbf{IV} (search space restrictions). This can be explained as \textbf{I} and \textbf{II} are DARTS' most prominent issues and thus constitute the main leads to pursue any follow-up work. On the other hand, non-DARTS-based DNAS saw those issues are inherent to DARTS and proposed a change of paradigm that allowed them to focus on other, more global, problems (i.e., \textbf{III} and \textbf{IV}).
Let us dive into the main conclusions of each category.

\textbf{(I)} A large subset of methods \cite{wang2021rethinking, chu2020fair, gu2021dots, yang2021towards, yu2022cyclic, cai2018proxylessnas} agreed that the final discretized architecture is dissimilar to the proxy model used during the search process, thus resulting in the optimization gap. However, these works differ in their proposed solutions to that analysis. Some replaced the discretization process to yield models that better fit the proxy network, while others designed a novel search process when the proxy network is more closely tied to the final model (or even a proxyless search process \cite{cai2018proxylessnas}). These DNAS works yielded improved results that show the relevance of their respective contributions. However, these improvements are often marginal (i.e., less than 1 \% top 1 accuracy improvement on ImageNet \cite{deng2009imagenet} compared to previous state-of-the-art), and there is no general consensus among all recent articles on how to reduce the optimization gap. Thus, this may indicate that, despite efforts to provide mathematical background, we still lack a formal model that would bring an optimal solution to this problem. The DNAS optimization gap is not closed yet.

\textbf{(II)} One interesting fact to note is that nearly every paper that addresses the \textit{skip connection} issue provides an analysis of why DARTS fails and draws similar conclusions: the non-parametric operations have an unfair advantage as they accelerate gradient descent in the early stage by forming structures akin to residual blocks \cite{he2016deep}. Eventually, this unfair competition suppresses parametric operations and leads to architectural \say{overfitting}. Furthermore, most of the reviewed works (e.g., \cite{chu2020fair, ye2022b, chen2019progressive}) proposed to add regularization on the search space to prevent this phenomenon. This regularization is generally applied either before ($\alpha$ weights) or after ($\beta$ weights) the \textit{softmax} relaxation. This proved relevant as adding regularization prevented \textit{skip connections} from becoming dominant and improved performance. This outcome is logical as a search space mixing parametric and non-parametric operations is inherently unbalanced, and regularization is a well-explored solution to overfitting and rebalancing ill-formed problems \cite{bishop2006pattern, tibshirani1996regression}. Finally, as an alternative solution, other works \cite{liang2019darts+, Zela2020Understanding} devised early-stopping mechanisms to stop the search process before the architecture overfits. However, these approaches are based on arbitrary or empirical criteria that are less formal than regularization-based approaches, hence explaining the popularity of the latter.

\textbf{(III)} Some approaches \cite{xu2019pc, yan2022radars, wang2021learning} managed to reduce the search cost drastically (up to 43 times for VIM-NAS \cite{wang2021learning}) compared to DARTS.  This made it possible to launch the search process on low-end, consumer-grade GPUs (and even CPUs in some cases). Hence, it contributed to making DNAS a very accessible process to automate neural network design. However, other methods \cite{wan2020fbnetv2, cai2018proxylessnas, nayman2021hardcore} chose to trade computational cost for reduced latency at inference, hence helping Deep Learning to deploy on low-resource devices, such as mobile phones. They did so by adding the inference latency as a differentiable objective so that raw performance is no longer the only goal of the search process. To save computational resources, the latency values for a specific platform are often retrieved from a latency lookup table.

\textbf{(IV)} As previously stated, the fact that most DNAS methods that addressed search space restrictions are not DARTS-based highlights that it is an issue more closely associated with DARTS. Thus, those methods had to craft search algorithms and/or search spaces that deeply diverge from DARTS. Most notably, all approaches in that area abandoned the cell-based building block paradigm as it is one of the main elements that restrict the search space. For instance, the FBNet family \cite{wu2019fbnet, wan2020fbnetv2, dai2021fbnetv3, wu2021fbnetv5} relied on a novel MobileNet-like search space that is block-based rather than cell-based. ProxylessNAS \cite{cai2018proxylessnas}, and DenseNAS \cite{fang2020densely} also leveraged a similar search space. Finally, the low number of studies tackling the search restrictions (i.e., 6) means that researchers mostly focused on other issues judged more urgent. In addition, using a larger search space leads to a drastic increase in computational cost, hence making it necessary to design mechanisms to save resources. This fact might explain why proposing a novel method addressing this issue is difficult.

Overall, over the past few years, all of these works contributed to making DNAS more and more viable, with the ability to discover architectures that can now far surpass the performance of handcrafted ones (e.g., CDARTS \cite{yu2022cyclic} reaches 78.2 \% top 1 accuracy on ImageNet \cite{deng2009imagenet} vs. 74.7 \% for MobileNetV2 \cite{sandler2018mobilenetv2}). In addition, the computational cost (i.e., the number of FLOPs) and the latency can be restrained to deploy models on embedded platforms such as mobile phones \cite{cai2018proxylessnas, wu2019fbnet, nayman2021hardcore}. This makes DNAS (and Deep Learning by extension) easier to deploy to solve real-world tasks and accessible to a wider audience. Thus, one could argue that DNAS is now a maturing field, with some methods being included in major AutoML libraries such as Microsoft NNI \cite{nni2021} and NASLib \cite{naslib-2020}. 

Nevertheless, DNAS still suffers from limitations as none of the proposed could solve all of the four identified challenges at once. Hence, no DNAS method could impose itself as a novel standard. This may substantially explain how DARTS withstood the test of time and remains popular despite its age. In our opinion, this also indicates that DNAS has room for improvement and has not reached its full potential yet. 

As discussed in Section \ref{sec:applications}, DNAS has already been applied to a wide range of applications (mainly related to computer vision). In the near future, with DNAS becoming more and more robust, we can expect it to be applied to an ever-increasing number of fields. For instance, works on transformers improvements \cite{chitty2022neural} and self-supervised learning \cite{heuillet2023nasiam} have already started to emerge. Other already explored fields, such as Generative Adversarial Networks (GANs) design \cite{gao2020adversarialnas}, could be further expanded to other applications, such as face generation \cite{kammoun2022generative}.

Additionally, Explainable AI (XAI) is paramount to ensure the deployment of Deep Learning in everyday tasks. In the case of automated decision making (e.g., judicial case analysis or autonomous driving), people affected by those decisions would have trouble accepting them unless the model is able to provide argumentation \cite{arrieta2020explainable}. Furthermore, some governmental entities such as the European Union plan to require by law that AI models deployed on the market should be explainable \cite{ebers2020regulating}. Concerning NAS, only a few approaches have been proposed to directly search for explainable CNN models. However, most of the existing studies leverage evolutionary-based \cite{carmichael2021explainable, agiollo2021shallow2deep} or one-shot probabilistic NAS \cite{liu2021fox, nayman2022bilinear}. To the extent of our knowledge, none targeted DNAS specifically. Nonetheless, one method dubbed SNAS (Saliency-Aware NAS) \cite{hosseinisaliency} is noteworthy as it is post-hoc (i.e., agnostic of the backbone) and can compute saliency maps for a wide range of NAS algorithms, including DNAS methods such as DARTS \cite{liu2018darts}. Thus, Explainable NAS (XNAS) is still an emerging field, and we predict that it will expand greatly in the near future when DNAS methods overcome all their current challenges.

Finally, novel trends have recently appeared in the NAS landscape, with promising non-DNAS approaches \cite{mellor2021neural, chen2021neural}. We also witnessed the revival of previously explored concepts such as Reinforcement Learning or Evolutionary Algorithms \cite{dai2021fbnetv3}. Consequently, we can expect that, in the near future, DNAS will benefit from knowledge/experience transfer in NAS methods that could combine the strengths of several approaches.

\section{Conclusion}
\label{sec:conclusion}

In this survey, we presented a comprehensive review of 26 recent Differentiable Neural Architecture approaches. We identified 4 different challenges that DNAS faces and used them as a taxonomy to classify the methods we reviewed. We analyzed their respective results and concluded that differentiable methods have drastically progressed in the past few years. However, the proposed solutions lack consensus, and none could impose itself as a new NAS standard like DARTS (which clearly passed the test of time). Nonetheless, we expect DNAS to grow further in the near future and benefit from externalities due to the alternative NAS branches and novel trends. Thus, we hope that this survey will prove useful to orient future research on DNAS.


\bibliographystyle{ACM-Reference-Format}
\bibliography{sample-base}










\end{document}